%% file: main.tex
\documentclass{article}

\PassOptionsToPackage{numbers, compress}{natbib}

\usepackage[preprint]{neurips_2022}

\usepackage[utf8]{inputenc} 
\usepackage[T1]{fontenc}    
\usepackage{hyperref}       
\usepackage{url}            
\usepackage{booktabs}       
\usepackage{amsfonts}       
\usepackage{nicefrac}       
\usepackage{microtype}      
\usepackage{xcolor}         

\usepackage{amsmath}
\DeclareMathOperator*{\argminA}{arg\,min}

\DeclareMathOperator{\score}{score}
\DeclareMathOperator{\similarity}{sim}
\DeclareMathOperator{\source}{source}
\DeclareMathOperator{\target}{target}
\DeclareMathOperator{\dparray}{dp}

\usepackage{float}
\usepackage{subcaption}
\usepackage{listings}
\usepackage{diagbox}

\usepackage{algpseudocode}
\algnewcommand{\algorithmicand}{\textbf{ and }}
\algnewcommand{\algorithmicor}{\textbf{ or }}
\algnewcommand{\OR}{\algorithmicor}
\algnewcommand{\AND}{\algorithmicand}
\usepackage[linesnumbered,lined,boxed,commentsnumbered]{algorithm2e}

\usepackage{calc}
\usepackage{tikz}
\usepackage{tikz-qtree,tikz-qtree-compat}

\definecolor{cmap_black}{RGB}{0,0,0}
\definecolor{cmap_orange}{RGB}{230,159,0}
\definecolor{cmap_sky_blue}{RGB}{212,234,255}
\definecolor{cmap_bluish_green}{RGB}{0,158,115}
\definecolor{cmap_yellow}{RGB}{240,228,66}
\definecolor{cmap_blue}{RGB}{109,173,255}
\definecolor{cmap_vermillion}{RGB}{213,94,0}
\definecolor{cmap_reddish_purple}{RGB}{204,121,167}

\tikzstyle{stateTransition}=[-stealth, thick]
\usepackage{xinttools}

\usepackage{makecell}

\usepackage{amsthm}
\theoremstyle{definition}
\newtheorem{definition}{Definition}[section]

\def\algo{DPIAT} 

\usepackage{wrapfig}

\usepackage{setspace}

\SetAlCapSkip{0.2cm}

\usepackage{tocloft}
\usepackage{minitoc}

\newcommand\tabcap[1]{\small \textbf{\textsc{#1}}}

\usetikzlibrary{backgrounds} 
\usetikzlibrary{fit}
\usetikzlibrary{calc}

\usepackage{soul}

\usepackage{pifont}
\newcommand{\cmark}{\textcolor{green!50!black}{\ding{51}}}
\newcommand{\xmark}{\textcolor{red}{\ding{55}}}

\usepackage[customcolors]{hf-tikz}

\usepackage{xargs}
\usepackage[colorinlistoftodos,prependcaption,textsize=tiny]{todonotes}
\newcommandx{\change}[2][1=]{\todo[linecolor=blue,backgroundcolor=blue!25,bordercolor=blue,#1]{#2}}

\title{Breaking the Architecture Barrier: A Method for Efficient Knowledge Transfer Across Networks}

\author{\hspace{-5pt}Maciej A.~Czyzewski$^{1,\dagger}$
	\quad
	Daniel Nowak$^{1,\dagger}$%
	\quad
	Kamil Piechowiak$^{1,\dagger}$%
	\vspace{5pt} 
	\\
	$^1$ Poznan University of Technology \:\:\:
	$^\dagger$equal contrib. \\
}

\begin{document}


\doparttoc
\faketableofcontents 
\parttoc
	
\maketitle

\input{src/0_abstract_v4}

\input{src/1_intro}

\input{src/2_related_work}

\input{src/3_method}

\input{src/4_experiments}
\input{src/5_conclusions}

\input{src/6_references}


\input{src/8_appendix}
\end{document}

%% file: src/0_abstract_v4.tex
\begin{abstract}
Transfer learning is a popular technique for improving the performance of neural networks. However, existing methods are limited to transferring parameters between networks with same architectures. We present a method for transferring parameters between neural networks with different architectures. Our method, called \algo{}, uses dynamic programming to match blocks and layers between architectures and transfer parameters efficiently. Compared to existing parameter prediction and random initialization methods, it significantly improves training efficiency and validation accuracy. In experiments on ImageNet, our method improved validation accuracy by an average of 1.6 times after 50 epochs of training. \algo{} allows both researchers and neural architecture search systems to modify trained networks and reuse knowledge, avoiding the need for retraining from scratch. We also introduce a network architecture similarity measure, enabling users to choose the best source network without any training.
\end{abstract}


%% file: src/1_intro.tex

\section{Introduction}

Training multiple neural networks is common in real-world deep learning workflows. During research, minor revisions to the architecture are often made on an iterative basis to improve the model's performance. This is done regardless of the level of automation of the research workflow. But the consecutive models are often similar and may reach similar local optima when given the slightly transformed parameters of the previous model. Intuitively, these local optima are often the same or better \cite{gpipe, goingdeeper, Radford2019LanguageMA} if the expressivity of the network is increased and the same or worse in the opposite case \cite{lotteryticket}. Currently, there is no algorithm that focuses solely on transferring parameters between source and target networks without arbitrary architecture restrictions.

We propose \algo{}, a fast, training-free, dynamic programming method for transferring knowledge between source and target convolutional neural networks (CNNs) of different architectures (schematic overview in Figure \ref{fig:concept}). This technique does not require additional training for the transfer itself and provides faster convergence, making it a superior alternative to random initialization when a pretrained, similar network is available. With \algo{}, models retain some of their previous knowledge and continue to converge. While \algo{} can be used in any fully automatic setup, such as neural architecture search, the primary focus of this paper is on the generic situation where both the source and target architectures are known. This makes \algo{} a potential solution for reducing the computational expense of multiple networks training, and can accelerate research workflows with limited resources. Essentially, \algo{} enables faster training by making knowledge reusable between different network architectures. However, as the differences between architectures increase, knowledge is lost. We show that target networks, after transferring knowledge from source networks, converge faster than networks initialized with Xavier \cite{xavier2010} or Kaiming \cite{he2015delving} methods. Additionally, we show that source-target similarity and convergence rate are positively correlated.








\textbf{Our contribution.}
In this paper, we present a method for transferring parameters from a pre-trained network to a network with a novel architecture. Our focus is on networks with different structures or architectures. Our main contributions are:
\begin{enumerate}
    \item We introduce the problem of zero-shot network adaptation (IAT) and provide a detailed discussion of its challenges (Section \ref{sec:problem}).
    \item We propose the \algo{} algorithm, a dynamic programming approach for fast IAT (Section \ref{sec:method}), and discuss its limitations (Appendix \ref{appendix:limits}).
    \item We introduce a measure of network architecture similarity (Section \ref{sec:sim}, \ref{sec:expsim}).
    \item We demonstrate that our approach outperforms alternative techniques in experiments on multiple common benchmark datasets (CIFAR-100, FOOD101, and ImageNet) and in practically relevant scenarios (Section \ref{sec:scenarios}, \ref{sec:experiments}).
\end{enumerate}



    


\subsection{Problem definition}
\label{sec:problem}

We define IAT as the transfer of parameters between networks with different architectures, without any limitations on branching or scaling.
Networks can be seen as architectures instantiated with parameters, representing a function $n(x) = a\left(x, \theta\right)$ where $a$ is the architecture and $\theta$ are the parameters.
An architecture $a$ defines the operations to be performed on the input and parameters in order to produce the output.

%
%

\begin{definition}[Inter-architecture knowledge transfer] 
IAT is a zero-shot transformation of parameters $\theta_s \in \Theta_s$ trained on the source architecture $a_s$ and the source problem domain $D_s$ that aims to minimize a loss function defined for the target domain $D_t$ and the target architecture $a_t$.
$$IAT: \Theta_s \longrightarrow \Theta_t$$
$$\argminA_{\theta} \sum_{(x_i, y_i) \in D_t} L(y_i, a_t\left(x_i, \theta\right))$$
\end{definition}

\input{src/results/concept.tex}

\subsection{Example usage}
\label{sec:scenarios}


\textbf{Initialization problem.}
One common scenario where IAT can be used is when a novel network $N_2$ is applied to a new problem described by $D_2$. Until IAT, the only option was to train the network from scratch, which can be very computationally intensive. With IAT, we can transfer knowledge from a pre-trained network $N_1$ on a different problem domain $D_1$ to the new network $N_2$ on the target domain $D_2$, effectively replacing the need for random initialization. This is shown as follows:
$N_1$-$D_1$ $\rightarrow$ $N_2$-$D_2$
This allows us to leverage the knowledge learned by $N_1$ on $D_1$ to improve the performance of $N_2$ on $D_2$.
To use our approach, the following steps should be taken:
\textbf{(a)} Prepare a pool of pre-trained models on ImageNet.
\textbf{(b)} Use our similarity score to find the network $N_1$ in the pool that is most similar to the target network $N_2$.
\textbf{(c)} Use DPIAT to transfer weights from $N_1$ to $N_2$.
\textbf{(d)} Train $N_2$ using the transferred weights, which should be faster than training from scratch (as shown in Table \ref{tab:time_to_acc_cifar100} for DPIAT).
This process allows us to take advantage of the knowledge learned by a pre-trained network to improve the performance of a novel network on a new problem without extensive training on the new problem.

\textbf{Iterative experimentation.} 
In this scenario, we want to find the best model for solving a problem described by $D_2$. This could apply to a Kaggle-like competition where many different networks are tested. Assume that $N_2$ has been trained for 100 epochs, and we want to test a network $N_3$ that operates at a higher resolution, has more layers, or has modified parameter tensor shapes. Our options are as follows:
\textbf{(a)} Initialize the network randomly.
\textbf{(b)} Match layers manually and use parameter remapping, but this is a time-consuming process, especially if it is repeated many times.
\textbf{(c)} Automate the process using IAT:
 Use DPIAT to transfer weights from $N_2$, which is pre-trained on $D_2$, to $N_3$ for $D_2$.
Then train $N_3$ using the transferred weights, which should be faster than training from scratch (as shown in Table \ref{tab:time_to_acc_cifar100} for DPIAT (fine-tuned)).
This allows us to quickly adapt a pre-trained network to a new network with a different architecture, without the need for extensive training of the new network.

%% file: src/results/concept.tex

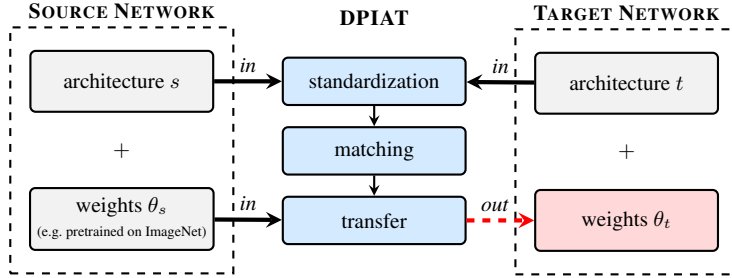
\begin{figure}[tbhp]
	\centering
	\small 
	\setlength{\tabcolsep}{2pt}
    

\begin{tikzpicture}[thick,scale=0.90, every node/.style={transform shape}]
 
\def \blockrowsep {1.1cm}

\def \groupixsep {2.5mm}
\def \groupiysep {3mm}

\def \groupheight {3cm}
\def \groupsep {1cm}

\def \maincolor {cmap_sky_blue} 
\def \outputcolor {cmap_yellow}

\tikzstyle{block}=[
    minimum width=2.7cm,
    minimum height=0.9cm, 
    draw=black,
    fill=black!5,
    rounded corners=2pt
]
        
\node[block, minimum width=2.7cm] (arch_a) {\small architecture $s$};  
\node[block, minimum width=2.7cm, below=\blockrowsep of arch_a] (w_a) {
    \makecell[c]{
        weights $\theta_s$ \\ \tiny{(e.g.\ pretrained on ImageNet)}
    }
};  

\node (a_plus) at ($(arch_a)!0.5!(w_a)$) {$+$};

\node[draw, dashed,
    inner xsep=\groupixsep,
    inner ysep=\groupiysep,
     fit=(w_a)(arch_a),
     minimum height=\groupheight,
     label={90:\tabcap{Source Network}}
     ](model_a){}; 
     
\node[block, minimum height=0.7cm, fill=\maincolor, right=\groupsep of arch_a] (iat_ast) {standardization};  
\node[block, minimum height=0.7cm, fill=\maincolor, below=0.3cm of iat_ast] (iat_match) {matching};  
\node[block, minimum height=0.7cm, fill=\maincolor, below=0.3cm of iat_match] (iat_transfer) {transfer};  

\node[draw=white,
    inner xsep=\groupixsep,
    inner ysep=\groupiysep, 
    fit=(iat_ast)(iat_match)(iat_transfer),
    minimum height=\groupheight,
     label={90:\tabcap{\algo{}}}
     ](iat){}; 
     
\draw[stateTransition, draw=black] (iat_ast) -- (iat_match);
\draw[stateTransition, draw=black] (iat_match) -- (iat_transfer);


\draw[stateTransition, draw=black, ultra thick] (arch_a) -- node[above] {\textit{in}} (iat_ast);
\draw[stateTransition, draw=black, ultra thick] (w_a) -- node[above] {\textit{in}} (iat_transfer);

\node[block, fill=red!15, right=\groupsep of iat_transfer] (w_b) {weights $\theta_t$};  
\node[block, above=\blockrowsep of w_b] (arch_b) {\small architecture $t$};  

\node (b_plus) at ($(arch_b)!0.5!(w_b)$) {$+$};

\node[draw, dashed,
    inner xsep=\groupixsep,
    inner ysep=\groupiysep,
     fit=(w_b)(arch_b),
    minimum height=\groupheight,
     label={90:\tabcap{Target Network}}
     ](model_b){}; 

\draw[stateTransition, draw=red, dashed, ultra thick] (iat_transfer) -- node[above] {\hspace{-0.5em}\textit{out}} (w_b);
\draw[stateTransition, draw=black, ultra thick] (arch_b) -- node[above] {\textit{in}} (iat_ast);






\end{tikzpicture}

\caption{
\small An overview of \algo{} (Section~\ref{sec:method}). IAT produces weights $\theta_t$ for the target network. The transfer operation includes the transformation and reassignment of the parameters.
}

\label{fig:concept}
\vspace{-3pt}
\end{figure}

%% file: src/2_related_work.tex
\vspace{-0.2cm}
\section{Related work}
\label{sec:relatedwork}

\vspace{-0.2cm}

\textbf{Random weight initialization.}
Various schemes were proposed to normalize the input and output layer variance, such as the Xavier initialization scheme \cite{xavier2010}. It has been shown that proper initialization can significantly improve the training of deep networks \cite{lecun1998}, while arbitrary initialization can slow down the optimization process \cite{mishkin2015}. Recent work has also focused on eliminating the need for normalization during training through the use of appropriate initialization schemes \cite{fixupinit, metainit}. Note that in most cases our IAT method is in fact competing with random initialization and
parameter prediction as these are the only available alternatives.




\textbf{Generality of layers' features.} An important factor in knowledge transfer is whether the parameters in the layers are universal enough to adapt quickly to new information flow.   
An analysis \cite{NIPS2014_375c7134} of the generality of features present in neural networks at different depths shows that layer order matters. Thus, we preserve the order of layers in \algo{}.







\textbf{Parameter Remapping.} IAT is often not named explicitly and applied in part in NAS.
To the best of our knowledge it has never been posed as a separate problem without the commonly assumed limitations including arbitrary network scaling and branching.
There are, however, some instances of the transfer operators we apply that are being used in other studies.
Note, that what we refer to as IAT has sometimes been limited and called parameter remapping. Nevertheless, the ``transfer'' nomenclature had been used originally in ``transfer learning'' \cite{NIPS2014_375c7134} and arguably should be used for other kinds of knowledge transfer between networks.



\textbf{(a) Net2Net.} 
This approach was first introduced in \cite{chen2016net2net}. It proposes a function-preserving transformation to transfer the weights of a network to a new, deeper (Net2DeeperNet) or wider (Net2WiderNet) network. However, the identity function cannot always be used (e.g. in the case of bottleneck layers), and the method only allows for the construction of deeper or layer-wise wider networks, ignoring network branching and skip-connections. Net2Net also does not allow for any branching of the network apart from predefined paths.
These transformations were later applied in NAS works, such as \cite{cai2017efficient} and \cite{cai2018pathlevel}. The path-level network transformations introduced in \cite{cai2018pathlevel} allow for the replacement of a single layer with a multi-branch pattern, but not vice versa. This approach allows for the construction of an equivalent N-branch motif using a convolutional (or other) layer, where the branches mimic the output of the original layer. However, this method only allows for one-way replacement and does not allow for the transfer of parameters from one multi-branch to another of a different type.


\textbf{(b) FNA.}
Several NAS studies \cite{pham2018enas, fang2019etnas, fang2020fast, fang2020fna++}, have investigated the use of parameter sharing through parameter remapping. This suggests that a wide range of architectures can effectively reuse the same set of parameters. In particular, FNA \cite{fang2020fast} and FNA++ \cite{fang2020fna++} extended the parameter remapping technique of Net2Net \cite{elsken2018efficient} by incorporating kernel-wise depth scaling. However, these approaches still suffer from the fundamental limitations of Net2Net, as described in Appendix \ref{appendix:faq-parameter-remapping}. Specifically, they rely on scaling of a network in an arbitrary manner.




\textbf{Parameter Prediction.} Prior studies have shown the parameter prediction to be feasible \cite{denil2013predicting, ha2017hypernetworks}. In particular, more recent work \cite{zhang2019ghn} based on graph neural networks showed promising results. In the most recent work on graph hypernetworks (GHN) \cite{fb2021ppuda}, authors create a network designed to learn to generate parameters for every architecture for a given dataset. Although this can be used to generate parameters for the target network which are then fine-tuned on a different domain, GHN is very resource intensive to train (reaching 50 days on 4xV10032GB for each 150 epochs on ImageNet). What is more, GHN parameter prediction used before further training is inferior to the random He initialization. This was concluded by the authors of the paper in the limitation's section at the bottom of the appendix. GHNs are incapable of providing a good initialization point for further fine-tuning which limits the use-cases of the method. These disadvantages prompt us to present the lightweight \algo{} with superior convergence. 

\textbf{Knowledge distillation.} It is a technique \cite{hinton2015, gou2020knowledge} that enables the transfer of knowledge from a larger network to a smaller one. However, this transfer requires the target network to be trained, which makes knowledge distillation unsuitable for many applications.



\textbf{Network Alignment.}
One of the main challenges related to IAT is developing an algorithm to solve the matching problem, i.e., selecting the source and target layers for transfer methods. This problem can be viewed as a network alignment problem, which involves identifying corresponding nodes in different networks. In this work, we present a fast DP algorithm to address this problem, but other approaches for network allignment have been proposed, such as REGAL \cite{heimann2018regal}, which uses the K-means algorithm to identify similar representations between two networks, and HashAlign \cite{heimann2018hashalign}, a hash-based framework that leverages structural, edge, and node properties and introduces new LSH techniques and graph features specifically tailored for network alignment. HashAlign expresses all alignments in terms of a center graph to avoid the all-pairwise-comparison problem.


%% file: src/3_method.tex

\section{Method}
\label{sec:method}

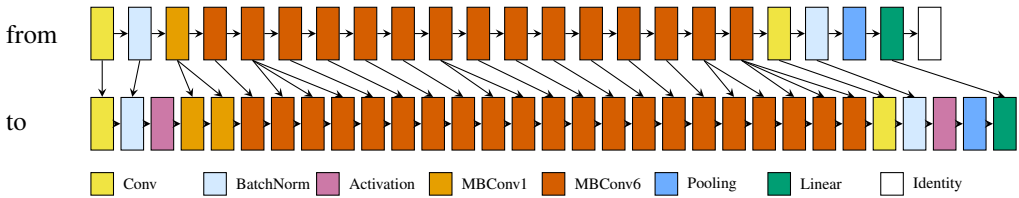
\begin{figure}[h]
    \centering
    \begin{tikzpicture}
        \def \distt {0.5cm}
        \def \dists {0.4cm}
        \def \sep {-1.2cm}
        \tikzstyle{nn_node}=[minimum width=0.3cm, minimum height=0.7cm, draw]
        
        \tikzstyle{mbconv1}=[fill=cmap_orange]        
        \tikzstyle{mbconv6}=[fill=cmap_vermillion]      
        \tikzstyle{conv}=[fill=cmap_yellow]        
        \tikzstyle{bn}=[fill=cmap_sky_blue]
        \tikzstyle{act}=[fill=cmap_reddish_purple]
        \tikzstyle{pool}=[fill=cmap_blue]     
        \tikzstyle{fc}=[fill=cmap_bluish_green]        
        \tikzstyle{identity}=[fill=white]
        \node[nn_node,conv,
        ] (te0) at (0*\distt, 0) {};
        \node[nn_node,bn] (te1) at (1*\distt, 0) {};
        \node[nn_node,mbconv1] (te2) at (2*\distt, 0) {};
        \node[nn_node,mbconv6] (te3) at (3*\distt, 0) {};
        \node[nn_node,mbconv6] (te4) at (4*\distt, 0) {};
        \node[nn_node,mbconv6] (te5) at (5*\distt, 0) {};
        \node[nn_node,mbconv6] (te6) at (6*\distt, 0) {};
        \node[nn_node,mbconv6] (te7) at (7*\distt, 0) {};
        \node[nn_node,mbconv6] (te8) at (8*\distt, 0) {};
        \node[nn_node,mbconv6] (te9) at (9*\distt, 0) {};
        \node[nn_node,mbconv6] (te10) at (10*\distt, 0) {};
        \node[nn_node,mbconv6] (te11) at (11*\distt, 0) {};
        \node[nn_node,mbconv6] (te12) at (12*\distt, 0) {};
        \node[nn_node,mbconv6] (te13) at (13*\distt, 0) {};
        \node[nn_node,mbconv6] (te14) at (14*\distt, 0) {};
        \node[nn_node,mbconv6] (te15) at (15*\distt, 0) {};
        \node[nn_node,mbconv6] (te16) at (16*\distt, 0) {};
        \node[nn_node,mbconv6] (te17) at (17*\distt, 0) {};
        \node[nn_node,conv] (te18) at (18*\distt, 0) {};
        \node[nn_node,bn] (te19) at (19*\distt, 0) {};
        \node[nn_node,pool] (te20) at (20*\distt, 0) {};
        \node[nn_node,fc] (te21) at (21*\distt, 0) {};
        \node[nn_node,identity] (te22) at (22*\distt, 0) {};

        \foreach \i in {1,...,22}
            \pgfmathsetmacro{\iprev}{\i-1}
            \draw[-stealth] (te\iprev) -- (te\i.west);
        
        \node[nn_node,conv, 
        ] (st0) at (0*\dists, \sep) {};
        \node[nn_node,bn] (st1) at (1*\dists, \sep) {};
        \node[nn_node,act] (st2) at (2*\dists, \sep) {};
        \node[nn_node,mbconv1] (st3) at (3*\dists, \sep) {};
        \node[nn_node,mbconv1] (st4) at (4*\dists, \sep) {};
        \node[nn_node,mbconv6] (st5) at (5*\dists, \sep) {};
        \node[nn_node,mbconv6] (st6) at (6*\dists, \sep) {};
        \node[nn_node,mbconv6] (st7) at (7*\dists, \sep) {};
        \node[nn_node,mbconv6] (st8) at (8*\dists, \sep) {};
        \node[nn_node,mbconv6] (st9) at (9*\dists, \sep) {};
        \node[nn_node,mbconv6] (st10) at (10*\dists, \sep) {};
        \node[nn_node,mbconv6] (st11) at (11*\dists, \sep) {};
        \node[nn_node,mbconv6] (st12) at (12*\dists, \sep) {};
        \node[nn_node,mbconv6] (st13) at (13*\dists, \sep) {};
        \node[nn_node,mbconv6] (st14) at (14*\dists, \sep) {};
        \node[nn_node,mbconv6] (st15) at (15*\dists, \sep) {};
        \node[nn_node,mbconv6] (st16) at (16*\dists, \sep) {};
        \node[nn_node,mbconv6] (st17) at (17*\dists, \sep) {};
        \node[nn_node,mbconv6] (st18) at (18*\dists, \sep) {};
        \node[nn_node,mbconv6] (st19) at (19*\dists, \sep) {};
        \node[nn_node,mbconv6] (st20) at (20*\dists, \sep) {};
        \node[nn_node,mbconv6] (st21) at (21*\dists, \sep) {};
        \node[nn_node,mbconv6] (st22) at (22*\dists, \sep) {};
        \node[nn_node,mbconv6] (st23) at (23*\dists, \sep) {};
        \node[nn_node,mbconv6] (st24) at (24*\dists, \sep) {};
        \node[nn_node,mbconv6] (st25) at (25*\dists, \sep) {};
        \node[nn_node,conv] (st26) at (26*\dists, \sep) {};
        \node[nn_node,bn] (st27) at (27*\dists, \sep) {};
        \node[nn_node,act] (st28) at (28*\dists, \sep) {};
        \node[nn_node,pool] (st29) at (29*\dists, \sep) {};
        \node[nn_node,fc] (st30) at (30*\dists, \sep) {};

        \foreach \i in {1,...,30}
            \pgfmathsetmacro{\iprev}{\i-1}
            \draw[-stealth] (st\iprev) -- (st\i.west);
            
        \draw[-stealth] (te0.south) -- (st0.north);
        \draw[-stealth] (te1.south) -- (st1.north);
        \draw[-stealth] (te2.south) -- (st3.north);
        \draw[-stealth] (te2.south) -- (st4.north);
        \draw[-stealth] (te3.south) -- (st5.north);
        \draw[-stealth] (te4.south) -- (st6.north);
        \draw[-stealth] (te4.south) -- (st7.north);
        \draw[-stealth] (te4.south) -- (st8.north);
        \draw[-stealth] (te5.south) -- (st9.north);
        \draw[-stealth] (te6.south) -- (st10.north);
        \draw[-stealth] (te7.south) -- (st11.north);
        \draw[-stealth] (te8.south) -- (st12.north);
        \draw[-stealth] (te9.south) -- (st13.north);
        \draw[-stealth] (te9.south) -- (st14.north);
        \draw[-stealth] (te10.south) -- (st15.north);
        \draw[-stealth] (te11.south) -- (st16.north);
        \draw[-stealth] (te12.south) -- (st17.north);
        \draw[-stealth] (te13.south) -- (st18.north);
        \draw[-stealth] (te14.south) -- (st19.north);
        \draw[-stealth] (te15.south) -- (st20.north);
        \draw[-stealth] (te16.south) -- (st21.north);
        \draw[-stealth] (te16.south) -- (st22.north);
        \draw[-stealth] (te17.south) -- (st23.north);
        \draw[-stealth] (te17.south) -- (st24.north);
        \draw[-stealth] (te17.south) -- (st25.north);
        \draw[-stealth] (te18.south) -- (st26.north);
        \draw[-stealth] (te19.south) -- (st27.north);
        \draw[-stealth] (te21.south) -- (st30.north);
        
        \tikzstyle{legend_box}=[minimum width=0.3cm, minimum height=0.3cm, draw]
        \def \legendsep {-2cm}
        \def \legenddist {1.5cm}
        \node[legend_box,conv] (l0) at (0*\legenddist, \legendsep) {};
        \node[legend_box,bn] (l1) at (1*\legenddist, \legendsep) {};
        \node[legend_box,act] (l2) at (2*\legenddist, \legendsep) {};
        \node[legend_box,mbconv1] (l3) at (3*\legenddist, \legendsep) {};
        \node[legend_box,mbconv6] (l4) at (4*\legenddist, \legendsep) {};
        \node[legend_box,pool] (l5) at (5*\legenddist, \legendsep) {};
        \node[legend_box,fc] (l6) at (6*\legenddist, \legendsep) {};
        \node[legend_box,identity] (l7) at (7*\legenddist, \legendsep) {};
        \node[right] at (l0.east) {\tiny Conv};
        \node[right] at (l1.east) {\tiny BatchNorm};
        \node[right] at (l2.east) {\tiny Activation};
        \node[right] at (l3.east) {\tiny MBConv1};
        \node[right] at (l4.east) {\tiny MBConv6};
        \node[right] at (l5.east) {\tiny Pooling};
        \node[right] at (l6.east) {\tiny Linear};
        \node[right] at (l7.east) {\tiny Identity};

        \node[
            below=0 of te0,
            xshift=-0.75cm, yshift=0.6cm,
        ]{\parbox{\widthof{another}}{from}};
        \node[
            below=0 of st0,
            xshift=-0.75cm, yshift=0.6cm,
        ]{\parbox{\widthof{another}}{to}};
    \end{tikzpicture}
    \caption{\small 
    The process of matching between a standardized ReXNet (x1.0) (top) and a standardized EfficientNet-B2 (bottom). The rectangles represent individual layers or blocks (such as MBConv \cite{mobilenetv2}). Layers within matched compound blocks are also matched. Low-level matching occurs between each parameter tensor within the block.
    }
    \label{example_matching}
    
\vspace{-3pt}
\end{figure}

The goal of IAT is to transfer parameters from a pre-trained network (the source) to a new network (the target) in order to have a better starting point for learning. 
A general framework for IAT can be described as a series of sequential steps: \textit{standardization} $\rightarrow$ \textit{matching} $\rightarrow$ \textit{transfer}. 
First, the standardization step provides consistent representations of the networks.
A scoring function assesses the similarity of source-target layer pairs. Then, the matching algorithm pairs up the source and target layers. Finally, the transfer method moves weights between layers.
Below we present variants of each step that form the best inter-architecture transfer method. The results of our ablation studies, in which we evaluated other combinations, can be found in Appendix \ref{appendix:ablation}.
The order of the DPIAT is $O(n^2m + p_s + p_t)$ where n is the number of layers in the target and m is the number of layers in the source, $p_s$ is the number of parameters in the source and $p_t$ is the number of parameters in the target. A detailed analysis of the time complexity can be found in Appendix \ref{appendix:complexity}.



\subsection{Standardization}

To process diverse architectures, a consistent representation must be created. This allows different architectures to be compared and their layers matched.
Consequently, the first step is to standardize the networks into building blocks that represent complex abstract notions that are commonly encountered. 
A block is defined as a group of layers that might follow different paths and are joined together through splits and merges.
In a ResNet architecture, \cite{kaiming2014resnet} a single building block defined by authors is composed of: \textbf{(a)} 2 convolutional layers, \textbf{(b)} 2 batch normalization layers, \textbf{(c)} 2 non-linear activation functions, \textbf{(d)} a shortcut connection. 

\algo{} divides a network into such blocks.
A network is represented as a graph of blocks,  with each block represented as a list of layers. 
This two-level approach provides a more structural representation of the information flow.
Figure \ref{example_matching} shows a simplified example of standardization and block matching.
Single convolutional and batch norm layers from the beginnings and ends of the networks were extracted.
In the middle, it correctly classified inverted residual bottlenecks as separate entities.

The standardization relies on the structure of a network defined by its author in its implementation. Networks consist of modules which contain submodules and so on. The structure of a network implementation is a tree. Leaves of the tree are single layers. \algo{} traverses this tree using DFS. In each node it makes a decision if all children should be merged into a single block or should be retained as separate blocks. Detailed overview and pseudo-code is available in the Appendix \ref{appendix:standarization}.


\subsection{Scoring}


The matching algorithm uses a similarity scoring function to pair layers. The similarity measure is based on the shapes of the underlying weight tensors.
We use a shape-based tensor similarity metric.
To receive a non-zero score, pairs of tensors must belong to layers of the same type (e.g.\ both are weights of a 2d convolutional layer).
This also ensures the same number of dimensions.
Let $S$ and $T$ be the respective sizes of source and target tensors that are being compared ($S$'s size: $s_1 \times s_2 \times ... \times s_n$, $T$'s size: $t_1 \times t_2 \times ... \times t_n$). Then we define the score as:
\begin{equation}\label{e:score}
    score(T, S) = \prod_{i=1}^{n} \min (\frac{t_i}{s_i}, \frac{s_i}{t_i})
\end{equation}
For every dimension the score function computes:
\begin{itemize}
    \item[(a)] what part of the source's dimension can be transferred to the target's dimension (when $s_i \ge t_i$)
    \item[(b)] what part of the target's dimension can be filled with unique parameters from the source's dimension (when $s_i \le t_i$).
\end{itemize}
The final score is the product of dimensions' ratios.
When all dimensions of a target tensor are smaller than source tensor dimensions, the score is equal to the percentage of transferred parameters from the source tensor. When all target dimensions are greater than source tensor dimensions, the score is equal to the percentage of the newly filled part of the tensor.
The general case is the composition of these two.

\subsection{Matching}
The matching method receives the standardized networks as an input. Its objective is to generate pairs of source and target layers in such a way that the performance of transfer is high.
We could use just a shape based score to match layers, but this would lead to the loss of layers' order. Layers at different depths of the network have different responsibilities which are not permutable as concluded in \cite{NIPS2014_375c7134}. 
\algo{} preserves the ordering of layers. If we represent both networks as directed path graphs of layers and matchings as edges between those two paths, the edges should not intersect. 

The matching algorithm is based on dynamic programming. It was inspired by the dynamic programming algorithm for finding the longest common subsequence \cite{lcm_dynamic}.
In our case, the counterpart of a sequence length is the sum of similarities of matched elements.
The algorithm maximizes this sum in two separate phases.
\
In the first phase, the similarity of every pair of blocks is computed.
In the second phase, using the similarities between blocks, the similarity of the networks is determined and the blocks are matched.
In both phases, networks are represented as sequences and for every pair in source and target sequences, a score is available. 
In the first phase, the sequences consist of layers present in single blocks and their score is computed with a shape score function (Equation \ref{e:score}). 
In the second phase, sequences consist of blocks and the score for every pair is known from the previous phase.

In both phases, matching is performed using dynamic programming. In phase two, cell $(i, j)$ represents the matching of the first $i$ blocks of the target network with the first $j$ blocks of the source network. The matching can be created in several ways:
\begin{itemize}
    \item[(a)] no target block is assigned to $j$-th source block, matching is the same as at $(i, j-1)$.
    \item[(b)] no source block is assigned to $i$-th target block, matching is the same as at $(i-1, j)$.
    \item[(c)] the source $j$-th block is matched with every suitable target's block in the range $[k, i]$ and the prefixes of the target of length $k-1$ and the source of length $j-1$ are matched.
\end{itemize}
The exact equations (Eq. {\ref{e:dp1}}, \ref{e:dp2}) can be seen in Figure \ref{DPMatching}. In phase one, blocks are replaced with layers and networks are replaced with blocks.


When layer $j$ is assigned to more than one layer, scores are summed. To penalize long matches, the sum is divided by the square root of the length of the match. This encourages the algorithm to create more diverse matching because long matches are penalized. 
For instance, let's consider a situation when a part of the source consists of two blocks $s_1, s_2$ and a part of the target consists of three blocks $t_1, t_2, t_3$ and $score(t_i, s_j) = a > 0$ for every $i=1,2,3, j=1,2$. Without division all three matchings $[(s_1, t_1), (s_1, t_2), (s_1, t_3)]$, $[(s_2, t_1), (s_2, t_2), (s_2, t_3)]$, $[(s_1, t_1), (s_1, t_2), (s_2, t_3)]$ receive the same score (equal to $3a$) while only the third one uses weights from both source layers. To prevent such situations all matchings with a single source layer have to be done at once and their sum has to be multiplied by some function $f(d)$ where $d$ is the number of matched layers with a single source layer. We want $f(d)$ to be a decreasing function - adding more layers should give a lower score per layer. We also want $d \cdot f(d)$ to be an increasing function - adding more layers should increase the total score. An example of such family of functions is $f(d) = d^{-x}$ for $ 0 < x < 1$. We have chosen arbitrarily $x=\frac{1}{2}$ that yields $f(d) = \frac{1}{\sqrt{d}}$.

At every step of computations, chosen actions are recorded.
The cell $dp[n,m]$ contains the score obtained from the matching of the networks, where $n,m$ are the numbers of blocks in the target and the source network, respectively.
Matching of blocks can be restored by following the actions backward, starting from cell $(n, m)$.
When the block pairs are known, it is possible to match layers within blocks using the information from the first phase of the algorithm.



\input{src/results/dp.tex}


\subsection{Transfer}
\algo{} concludes with the parameter transfer.
Parameters can be transferred from source to target tensors once the tensors are matched.
The transfer method has to determine which parameters to transfer when the size of the source dimension is larger than the size of the target dimension or choose how and which parts of the target tensor will be filled if the size of the source dimension is smaller than the size of the target dimension.
%
If a layer contains both weights and biases, the mapping between parameters is computed for weights and the same indices are used for biases.

\algo{} copies the center crop of $T$ to the center crop of $S$. For each dimension, it chooses a smaller size and crops the larger size, cutting off a beginning and an end of the tensor so that the sizes are equal. The exact formula can be found in the Appendix \ref{appendix:transfer}.

The transfer method has the ability to preserve the flow of information from the original network. Let's consider a matching of two consecutive layers from a source $T_1, T_2$ to two consecutive layers in a target $S_1, S_2$.
Then the same weight is used in the corresponding neuron of $S_2$ and $T_2$ to process output of the given neuron from the previous layer. It would not be the case, if we permuted the neurons of the first layer (by not preserving the order of inputs while copying from $T_2$ to $S_2$ or not preserving the order of outputs while copying from $T_1$ to $S_1$).


This property can also be observed in the following situation. Imagine that source and target are convolutional networks with the same number of layers that differ only by the number of channels so that every layer in the target does not have less channels than its counterpart in the source. If we transfer weights from the source to the target and zero out all other weights in the source, both networks will be realizations of exactly the same function.


\subsection{Similarity of architectures}
\label{sec:sim}
As a byproduct of our matching functions, we have created a measure of the similarity of architectures. While choosing the best match, we must score different matching possibilities, resulting in a numerical value representing the quality of the match. Every target layer is matched to at most one source layer, and the score of the single match is at most 1. Therefore, the maximal score of the match is equal to the number of layers in the target network. In other words, the maximal value of the match is equal to the target matching with itself. With this knowledge, we can define a similarity of architectures:
\begin{equation}\label{e:fpresim}
    f(a, b) = \score(a,b) / \score(a,a)
\end{equation}
The values of Eq. \ref{e:fpresim} function belong to $[0, 1]$. However, it is not symmetric. To make it symmetric we can take $f$ in both directions and average the results. We take a geometric mean to lower the value when $f(a,b)$ and $f(b,a)$ are substantially different (this is the case when one network is significantly bigger than the other). Finally, we define similarity between two networks $\source, \target$ as:
\begin{equation}\label{e:sim}
    \similarity(\source, \target) = \sqrt{f(\source, \target)f(\target, \source)}
\end{equation}

%% file: src/results/dp.tex

\begin{figure}[h]
\vspace{-3pt}
\small 
\setlength{\tabcolsep}{2pt}
\tikzstyle{every picture}+=[remember picture]
\begin{minipage}[r]{0.25\textwidth}
\begin{tikzpicture} 
    \xintFor* #1 in {\xintSeq[1]{1}{7}} \do {
        \xintFor* #2 in {\xintSeq[1]{1}{7}} \do {
            \tikzstyle{s#1#2}=[draw]
        }
    }

    \tikzstyle{s71}=[fill=black]
    \tikzstyle{s52}=[fill=cmap_blue]
    \tikzstyle{s53}=[fill=cmap_sky_blue]
    \tikzstyle{s54}=[fill=white] 
    \tikzstyle{s45}=[fill=cmap_yellow]
    \tikzstyle{s42}=[fill=cmap_sky_blue]
    
    \foreach \i in {1,...,7}
        \foreach \j in {1,...,6}
            \node[s\i\j, draw, thick, minimum width=0.4cm,minimum height=0.4cm] (s\i\j) at (\i*0.4, \j*0.4)  {};
            
    \node (st_begin) at (0,2.8) {};
    \node (s_end) at (0,0) {};
    \node (t_end) at (3.2,2.8) {};
    
    \node (t_name) at (1.5,3.1) {\textbf{source} blocks};
    \node[rotate=90] (t_name) at (-0.3,1.5) {\textbf{target} blocks};
    
    \draw[stateTransition] (st_begin) -- (s_end);
    \draw[stateTransition] (st_begin) -- (t_end);
    
    \draw[stateTransition, draw=cmap_orange] (s45) -- (s41);
    
    \node [right of=s52, node distance=1.2cm] (pos_i) {i};
    \node [below of=s53, node distance=1.3cm] (pos_j) {j};
    \node [right of=s55, node distance=1.2cm] (pos_k) {k};
\end{tikzpicture}
\end{minipage}
\begin{minipage}[l]{0.75\textwidth}\centering
\vspace{-0.5cm}


\begin{equation}\label{e:dp1}\begin{split}
\tikzmarkin[
    set fill color=cmap_blue,
    set border color=white
]{c1}(0.05,-0.25)(-0.05,0.40)
    \dparray[i, j]
\tikzmarkend{c1}
&=\max\Bigg(
    \tikzmarkin[
        set fill color=cmap_sky_blue,
        set border color=white
    ]{c2}(0.05,-0.25)(-0.05,0.40)
        \dparray[i, j-1]
    \tikzmarkend{c2},
    \tikzmarkin[
        set fill color=cmap_sky_blue,
        set border color=white
    ]{c3}(0.05,-0.25)(-0.05,0.40)
        \dparray[i-1, j]
    \tikzmarkend{c3},\\
    &\max_{\substack{0 < k \leq i}}
        \frac{1}{\sqrt{i-k+1}} \sum_{l=k}^{i} \score(t_{l}, s_{j})
    +%
    \tikzmarkin[
        set fill color=cmap_yellow,
        set border color=white
    ]{c4}(0.05,-0.25)(-0.05,0.40)
        \dparray[k-1, j-1]
    \tikzmarkend{c4}
\Bigg)
\end{split}\end{equation}

\begin{equation}\label{e:dp2}\begin{split}
\score(\target, \source)&=%
    \tikzmarkin[
        set fill color=black!20,
        set border color=white
    ]{c5}(0.05,-0.25)(-0.05,0.40)
        \dparray[n, m]
    \tikzmarkend{c5}
\end{split}\end{equation}


\end{minipage}

\caption{\small 
Dynamic Programming Matching of blocks (or layers) in two networks. The cell $(i, j)$ represents the matching of the first $i$ blocks in the target network with the first $j$ blocks in the source network. The vertical axis represents the target network while the horizontal axis represents the source. The $t_l$ denotes the $l$-th block of the target network, and the $s_j$ denotes the $j$-th block of the source network.
}
\label{DPMatching}
\vspace{-3pt}
\end{figure}

%

%% file: src/4_experiments.tex
\section{Experiments}
\label{sec:experiments}

The \algo{} algorithm has been tested on four datasets: CIFAR-10, CIFAR-100 \cite{cifar}, Food-101 \cite{food101}, and ImageNet \cite{imagenet}. On the first three datasets, we trained the source networks, performed transfer between every possible pair of available networks, and then trained the target networks using the weights obtained in the IAT process.

The models and their ImageNet weights used in this paper were provided by the timm library \cite{rw2019timm}. All trainings were performed using the Adam optimizer \cite{kingma_adam} with an initial learning rate of $4\cdot10^{-3}$ and a cosine annealing schedule with a final learning rate of $4\cdot10^{-4}$. The batch size was set to 128. Detailed running times and convergence speedup comparisons are included in Appendix \ref{appendix:speed-up}.

\input{src/results/cifar100}

\subsection{Random initialization}

In this experiment, we trained nine models initialized randomly (using the same initialization schemes as in the original papers introducing the models). The networks used were: EfficientNet-B0, B1, B2 \cite{tan2019efficientnet}, MixNet-M, L \cite{tan2019mixconv}, ReXNet (x1.0), (x1.5) \cite{rexnet}, MnasNet, and SeMnasNet (MnasNet with squeeze-and-excitation) \cite{tan2019mnasnet}. The models were trained for 100 epochs.

For each target model, we initialized it randomly (using the same initialization schemes as the source models) and then performed knowledge transfer. We report the test accuracies obtained after 4 epochs of training and compare them with the accuracies obtained after 4 epochs of training starting from random initialization. We define the effectiveness as the ratio of these values.

The effectiveness of training on CIFAR-100 is presented in Table \ref{tab:cifar100}. The ratios for the other datasets can be found in the Appendix \ref{appendix:more} and \ref{appendix:imagenetpretraining}.


\input{src/results/cifar100_ghn}

In all cases, the test accuracy obtained after using \algo{} and training for 4 epochs was better than the test accuracy after random initialization and 4 epochs of training. More than double improvement was observed in the transfer between similar networks.

The results can also be compared to Table \ref{tab:cifar100_ghn}, which shows the effectiveness of GHN2 \cite{fb2021ppuda}. GHN2 gives results that are worse or comparable to random initialization.



\subsection{Transfer on CIFAR100}

\begin{figure}[H]
\vspace{-9pt}
\small 
\setlength{\tabcolsep}{2pt}
    \centering
    \includegraphics[width=1.0\textwidth]{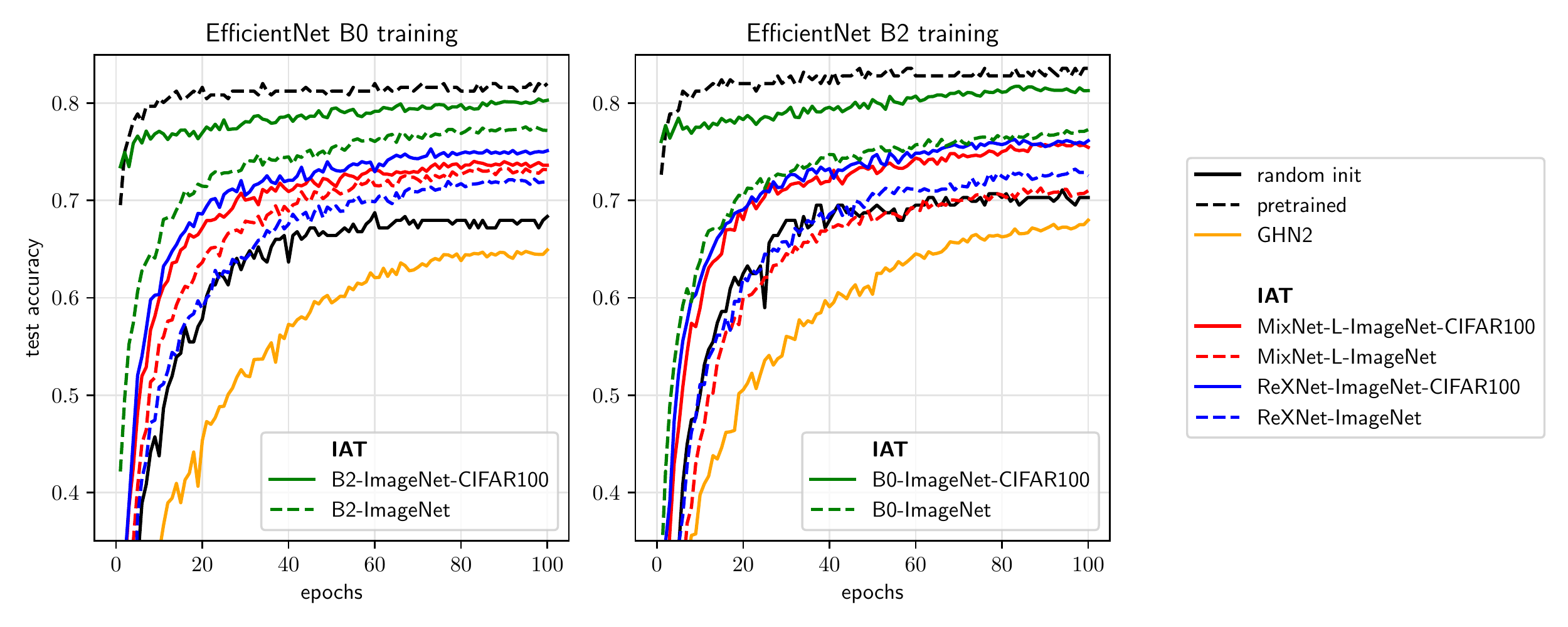}
    \caption{\small 
Test accuracy during CIFAR100 training. The black solid line shows the accuracy of a randomly initialized network, while the black dashed line shows the accuracy of a network initialized with ImageNet parameters. The yellow line shows the accuracy of a network initialized with GHN2. The other lines show the accuracy after using \algo{}. The "-ImageNet" suffix means that the source parameters are ImageNet pretrained, while the "-ImageNet-CIFAR100" suffix means that the source was initialized with ImageNet and fine-tuned on CIFAR100 for 100 epochs.
    }
    \label{fig:cifar100}
    \vspace{-3pt}
\end{figure}

Using \algo{}, we transferred parameters to EfficientNet-B0 from EfficientNet-B2, MixNet-L, and ReXNet, and to EfficientNet-B2 from EfficientNet-B0, MixNet-L, and ReXNet (see Figure \ref{fig:cifar100}). We considered 3 variants of each source network, containing ImageNet pretrained parameters (-ImageNet suffix), ImageNet pretrained parameters with 100 epochs of fine-tuning on CIFAR100 (-ImageNet-CIFAR100 suffix), and randomly initialized parameters with 100 epochs of training on CIFAR100 (-CIFAR100 suffix, see Figure \ref{fig:cifar100_random_init} for this variant). We then trained the target networks after IAT for 100 epochs on CIFAR100, as shown in Table \ref{tab:cifar100_long}. We compared our method with random initialization of the target models, parameters pretrained on the target models on ImageNet, and GHN2 \cite{fb2021ppuda} parameters predicted for the target models. Detailed plots of the total training on CIFAR100 (loss, train acc.) for \algo{}, random initialization, and GHN can be found in Appendix \ref{appendix:train-on-cifar100}.


In terms of test accuracy, the best results were obtained with ImageNet pretrained parameters. However, not all networks have such parameters available. In this case, it is beneficial to transfer parameters from a similar source network (such as transferring from EfficientNet-B0 to EfficientNet-B2). If the source network was not trained on the target domain but has ImageNet pretrained parameters, these parameters can still be used. Transferring these parameters from EfficientNet-B2 to B0 resulted in an accuracy of $0.776$ after 100 epochs. Transfers from less similar networks yielded worse results (MixNet-L - $0.733$, ReXNet - $0.722$), but still better than random initialization ($0.688$).


\input{src/results/cifar100_long}

In a scenario where many architectures are tested on a single domain, networks with ImageNet pretrained weights and then fine-tuned on the target domain parameters may be available. Transferring from such networks preserves some features learned during the initial fine-tuning. This results in even better performance on the target domain (EfficientNet-B2 - $0.804$, MixNet-L - $0.740$, ReXNet - $0.753$ when the target is EfficientNet-B0).


\subsection{Transfer on ImageNet}

We trained a randomly initialized EfficientNet-B0 on ImageNet and compared it with initialization using \algo{} from 3 different models (EfficientNet-B2, MixNet-L, ReXNet, see Figure \ref{fig:imagenet}) that were pretrained on ImageNet. We trained for 50 epochs, as detailed in Appendix \ref{appendix:train-on-imagenet}. Regardless of the source architecture used, the results were significantly better than in the case of random initialization. The best results were obtained with a transfer from EfficientNet-B2 (which has a similarity score of 0.8 with the target). After 50 epochs, the test accuracy was $0.699$ (compared to $0.624$ after 50 epochs with random initialization).



\begin{figure}[H]
\vspace{-3pt}
\small 
\setlength{\tabcolsep}{2pt}
\centering
\begin{minipage}[l]{0.5\textwidth}
    \centering 
    \includegraphics[width=0.80\textwidth]{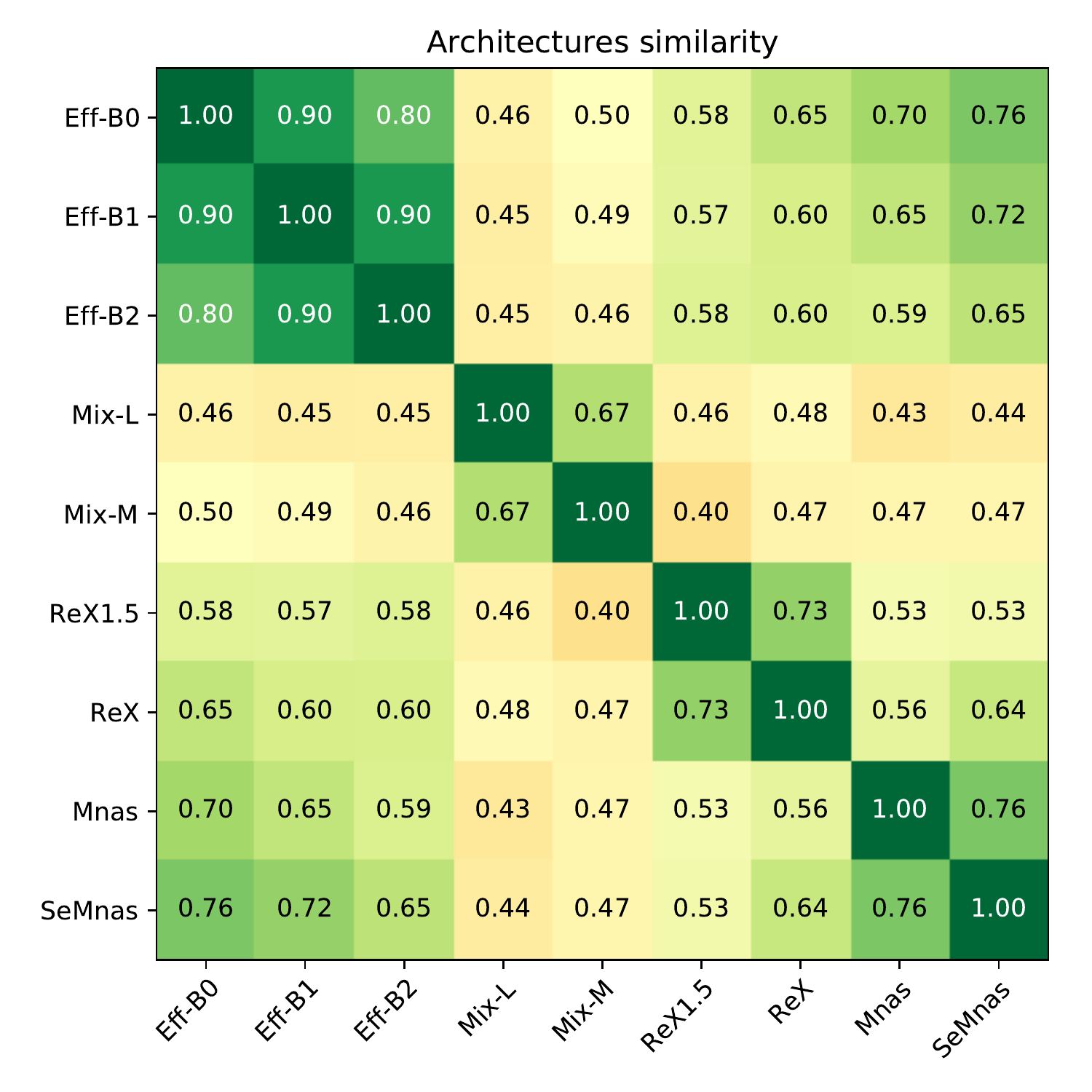}
    \captionof{figure}{
    \small Architectures similarity. Values close to 0 indicate that the architectures are significantly different, while values close to 1.0 indicate that they are identical. The method for calculating similarity is described in Section \ref{sec:sim} (see Eq. \ref{e:sim}).
    }
    \label{fig:sim}
    \vspace{-3pt}
\end{minipage}%
\hfill%
\begin{minipage}[r]{0.4\textwidth}
    \raggedleft 
    \includegraphics[height=5.5cm]{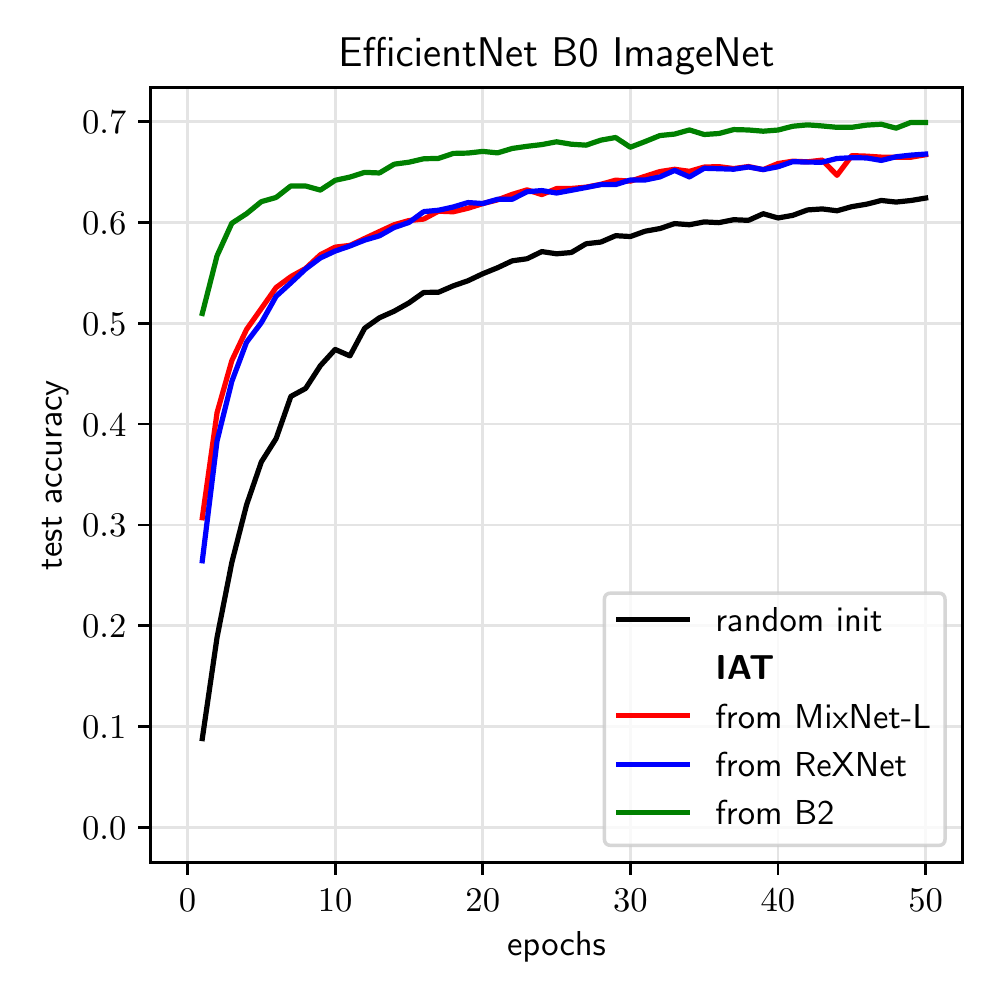}
    \captionof{figure}{
    \small Test accuracy over epochs during ImageNet training. The black solid line shows the accuracy of a randomly initialized network. The other lines show the accuracy after using \algo{} to transfer parameters from different source networks. 
    }
    \label{fig:imagenet}
    \vspace{-3pt}
\end{minipage}
\vspace{-3pt}
\end{figure}

\subsection{Similarity of architectures}
\label{sec:expsim}

As demonstrated in Section \ref{sec:method} and discussed in more detail in Appendix \ref{appendix:corr}, the scoring function used in \algo{} is shown to be correlated with the quality of IAT. As shown in Figure \ref{fig:sim} and \ref{fig:imagenet}, IAT between similar architectures is more effective. The low similarity between Eff-B0 and Mix-L is due to the branching strategy used in Mix-L. All EfficientNets belong to the same scalable family, making the transfer effective and the similarity measure high.

%% file: src/results/cifar100.tex
\begin{table}[tbhp] 
\centering
\vspace{-10pt}

\caption{\small
Effectiveness of knowledge transfer with \algo{} on CIFAR100 after 4 epochs of training. The table shows the sources in the top row and the targets in the first column. The standard deviations for the effectiveness are less than 0.1, with a median value of 0.02. The average effectiveness of random initialization is 1.0.
\looseness-1}
\vspace{0.2cm}

\scriptsize
\newcommand{\width}{0.135\textwidth}
\setlength{\tabcolsep}{3pt}

\begin{tabular*}{\textwidth}{@{\extracolsep{\fill}}r*{9}{c}}
\toprule

\multicolumn{1}{l}{}&\multicolumn{9}{c}{\tabcap{Source}} \\
\tabcap{Target} &    Eff-B0 &    Eff-B1 &    Eff-B2 &              Mix-L &              Mix-M &             ReX1.5 &                ReX &               Mnas &             SeMnas \\
\midrule
Eff-B0 &  $ 2.11 $ &  $ 2.02 $ &  $ 2.05 $ &           $ 1.42 $ &           $ 1.47 $ &           $ 1.51 $ &           $ 1.59 $ &  $ \mathbf{2.15} $ &           $ 2.13 $ \\
Eff-B1 &  $ 2.21 $ &  $ 2.15 $ &  $ 2.19 $ &           $ 1.46 $ &           $ 1.59 $ &           $ 1.58 $ &           $ 1.69 $ &           $ 2.20 $ &  $ \mathbf{2.22} $ \\
Eff-B2 &  $ 2.32 $ &  $ 2.29 $ &  $ 2.27 $ &           $ 1.58 $ &           $ 1.65 $ &           $ 1.71 $ &           $ 1.76 $ &           $ 2.23 $ &  $ \mathbf{2.33} $ \\
Mix-L  &  $ 1.34 $ &  $ 1.36 $ &  $ 1.41 $ &  $ \mathbf{1.99} $ &           $ 1.53 $ &           $ 1.32 $ &           $ 1.30 $ &           $ 1.60 $ &           $ 1.52 $ \\
Mix-M  &  $ 1.35 $ &  $ 1.40 $ &  $ 1.34 $ &           $ 1.46 $ &  $ \mathbf{1.92} $ &           $ 1.30 $ &           $ 1.22 $ &           $ 1.52 $ &           $ 1.58 $ \\
ReX1.5 &  $ 1.17 $ &  $ 1.09 $ &  $ 1.12 $ &           $ 1.08 $ &           $ 1.08 $ &  $ \mathbf{1.56} $ &           $ 1.22 $ &           $ 1.23 $ &           $ 1.29 $ \\
ReX    &  $ 1.24 $ &  $ 1.11 $ &  $ 1.16 $ &           $ 1.13 $ &           $ 1.11 $ &           $ 1.18 $ &  $ \mathbf{1.60} $ &           $ 1.28 $ &           $ 1.36 $ \\
Mnas   &  $ 1.98 $ &  $ 1.85 $ &  $ 1.80 $ &           $ 1.40 $ &           $ 1.46 $ &           $ 1.50 $ &           $ 1.48 $ &  $ \mathbf{2.20} $ &           $ 1.96 $ \\
SeMnas &  $ 2.04 $ &  $ 1.96 $ &  $ 1.95 $ &           $ 1.39 $ &           $ 1.50 $ &           $ 1.58 $ &           $ 1.44 $ &  $ \mathbf{2.19} $ &           $ 2.17 $ \\
\bottomrule
\end{tabular*}
\vspace{-3pt}
\vspace{-6pt}
\label{tab:cifar100}
\end{table}

%% file: src/results/cifar100_ghn.tex
\begin{wraptable}[10]{r}{7cm}
\centering

\caption{\small 
Effectiveness of initialization with GHN2, as described in \cite{fb2021ppuda}, after 4 epochs of training on CIFAR100. Values less than 1.0 indicate that it performs worse than random initialization (baseline).
\looseness-1}

\scriptsize
\newcommand{\width}{0.135\textwidth}
\setlength{\tabcolsep}{3pt}
    
\begin{tabular}{rccc}
    \toprule
    \multicolumn{1}{l}{}&\multicolumn{3}{c}{\tabcap{Initialization}} \\
    \tabcap{Target} &       GHN2-CIFAR10 & GHN2-ImageNet & random init \\
    \midrule
    Eff-B0 &  $ 0.80 $ & $ 0.72 $ & $ \mathbf{1.0} $ \\
    Eff-B1 &  $ 0.82 $ & $ 0.69 $ & $ \mathbf{1.0} $ \\
    Eff-B2 &  $ 0.81 $ & $ 0.66 $ & $ \mathbf{1.0} $ \\
    \bottomrule
\end{tabular}
\label{tab:cifar100_ghn}
\vspace{-3pt}
\end{wraptable}


%% file: src/results/cifar100_long.tex
\begin{table}[tbhp]
\centering
\vspace{-10pt}

\caption{\small Maximal test accuracy during 100 epochs of training on CIFAR100 for different initializations and source networks for \algo{}.\looseness-1}
\vspace{0.2cm}

\scriptsize
\newcommand{\width}{0.135\textwidth}
\setlength{\tabcolsep}{3pt}

\begin{tabular*}{\textwidth}{@{\extracolsep{\fill}}r*{3}c*{4}c*{4}c}
    \toprule
    \multicolumn{1}{l}{} &
        \multicolumn{3}{c}{} & 
        \multicolumn{8}{c}{\tabcap{Source}} \\
    \multicolumn{1}{l}{} &
        \multicolumn{3}{c}{\tabcap{Initialization}} &
        \multicolumn{4}{c}{\textbf{ImageNet-CIFAR100}} &
        \multicolumn{4}{c}{\textbf{ImageNet}} \\ 
    {\tabcap{Target}} & 
        random init & pretrained & GHN2 &
        Mix-L & ReX & Eff-B0 & Eff-B2 & 
        Mix-L & ReX & Eff-B0 & Eff-B2 \vspace{1pt}\\
\cline{1-4}\cline{5-8}\cline{9-12}\vspace{-4pt}\\
Eff-B0 &       0.688 &      \textbf{0.820} &  0.649 &  0.740 &  0.753 &                    - &                \textbf{0.804} &             0.733 &           0.722 &           - &       \textbf{0.776} \\
Eff-B2 &       0.711 &      \textbf{0.836} &  0.680 &  0.759 &  0.763 &                \textbf{0.817} &                    - &             0.713 &           0.732 &       \textbf{0.772} &           - \\
\bottomrule
\end{tabular*}
    \label{tab:cifar100_long}
    \vspace{-3pt}
    \vspace{-0.1cm} 
\end{table}

%% file: src/5_conclusions.tex
\section{Conclusions}
\label{sec:conclusions}






In this paper, we present a new direction for future research by defining inter-architecture knowledge transfer (IAT) and experimentally demonstrating its effectiveness in speeding up training. To the best of our knowledge, this is the first study on how to transfer parameters between two networks without arbitrary architecture constraints and training requirements.

We propose the \algo{} algorithm for IAT, which shows promising results. \algo{} is able to transfer parameters in a fraction of a second, making the knowledge reusable not only between domains but also between architectures. Networks initialized in this way train faster, showing properties superior to random and hypernetwork-based initialization. \algo{} does not suffer from exploding gradients in all the tested cases. While there is room for improvement in the way we align networks and in the overall performance of the method, \algo{} can already significantly reduce the computational burden needed for both automated neural architecture search and manual experimentation.

%% file: src/6_references.tex
\normalsize
\medskip
	
{\small
	\begin{spacing}{0.99}
		\bibliographystyle{unsrtnat}
		\bibliography{refs}
	\end{spacing}	
}

\vfill
	
\newpage
\normalsize

%% file: src/8_appendix.tex
\appendix
\setlength\cftbeforesubsecskip{-1pt}
\renewcommand\cftsecafterpnum{\vskip0pt}

\part{Appendix}\label{appendix}

\vspace{-10pt}
\begin{spacing}{0.2}
	\parttoc 
\end{spacing}

\section{\algo{} Details}

\subsection{Standardization} 
\label{appendix:standarization}

\begin{figure}[H]
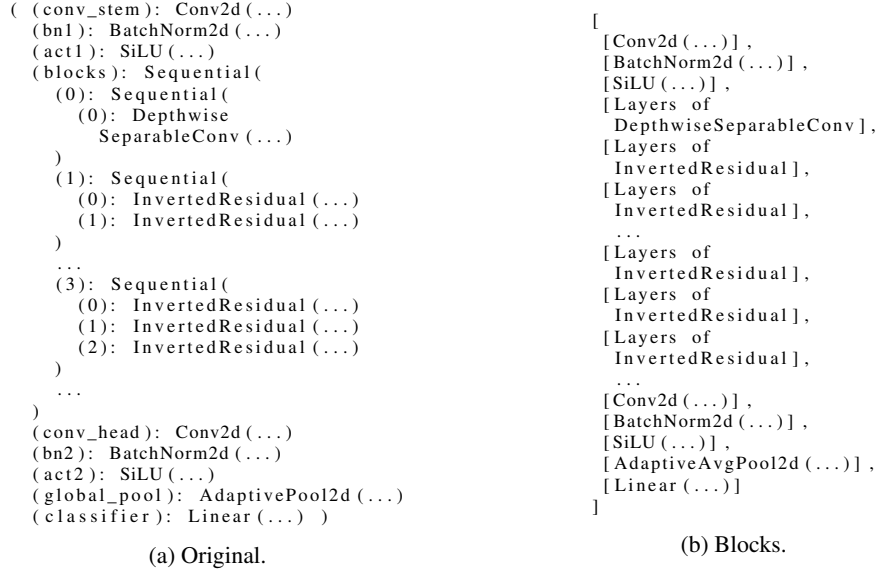

    \small 
\setlength{\tabcolsep}{2pt}
        \centering
        \begin{subfigure}[t]{.6\linewidth}
            \centering
            \begin{tabular}{c}
\begin{lstlisting}[basicstyle=\scriptsize]
( (conv_stem): Conv2d(...)
  (bn1): BatchNorm2d(...)
  (act1): SiLU(...)
  (blocks): Sequential(
    (0): Sequential(
      (0): Depthwise
        SeparableConv(...)
    )
    (1): Sequential(
      (0): InvertedResidual(...)
      (1): InvertedResidual(...)
    )
    ...
    (3): Sequential(
      (0): InvertedResidual(...)
      (1): InvertedResidual(...)
      (2): InvertedResidual(...)
    )
    ...
  )
  (conv_head): Conv2d(...)
  (bn2): BatchNorm2d(...)
  (act2): SiLU(...)
  (global_pool): AdaptivePool2d(...)
  (classifier): Linear(...) )
\end{lstlisting}
            \end{tabular}
        \caption{Original.}
        \end{subfigure}
                \begin{subfigure}[t]{.39\linewidth}
            \centering
            \begin{tabular}{c}
\begin{lstlisting}[basicstyle=\scriptsize]
[ 
 [Conv2d(...)],
 [BatchNorm2d(...)],
 [SiLU(...)],
 [Layers of
  DepthwiseSeparableConv],
 [Layers of
  InvertedResidual],
 [Layers of
  InvertedResidual],
  ...
 [Layers of
  InvertedResidual],
 [Layers of
  InvertedResidual],
 [Layers of
  InvertedResidual],
  ...
 [Conv2d(...)],
 [BatchNorm2d(...)],
 [SiLU(...)],
 [AdaptiveAvgPool2d(...)],
 [Linear(...)]
]
\end{lstlisting}
            \end{tabular}
        \caption{Blocks.}
        \end{subfigure}
        \caption{EfficientNet (a) transformed into a list of lists of layers (b).}
        \label{efficientnets-blocks-standardization}
        \vspace{-3pt}
    \end{figure}

Separating blocks is not always straightforward.
In some cases, there are small sub-blocks inside a block that have to be incorporated into their parent block. An example of such a situation can be seen in the implementation of the EfficientNet-B0 from the \textit{timm library} \cite{rw2019timm} (Figure \ref{efficientnets-blocks-standardization}).
Each Inverted Residual block contains a Squeeze-and-Excitation block. Basic building blocks of a network are Inverted Residual blocks, so Squeeze-and-Excitation block's layers have to be included in the representation of an Inverted Residual block (Figure \ref{inverted-residual-blocks-standardization}). Moreover, in this implementation Inverted Residual blocks are grouped. That increases the nesting level. The algorithm has to deal with such difficulties and correctly determine the basic building blocks.

    \begin{figure}[H]
    \small 
\setlength{\tabcolsep}{2pt}
        \centering
        \begin{subfigure}[t]{.5\linewidth}
            \centering
            \begin{tabular}{c}
\begin{lstlisting}[basicstyle=\scriptsize]
InvertedResidual(
  (conv_pw): Conv2d(...)
  (bn1): BatchNorm2d(...)
  (act1): SiLU(...)
  (conv_dw): Conv2d(...)
  (bn2): BatchNorm2d(...)
  (act2): SiLU(...)
  (se): SqueezeExcite(
    (conv_reduce): Conv2d(...)
    (act1): SiLU(...)
    (conv_expand): Conv2d(...) )
  (conv_pwl): Conv2d(...)
  (bn3): BatchNorm2d(...)
)
\end{lstlisting}
            \end{tabular}
        \caption{Original.}
        \end{subfigure}
                \begin{subfigure}[t]{.49\linewidth}
            \centering
            \begin{tabular}{c}
\begin{lstlisting}[basicstyle=\scriptsize]
[Conv2d(...),
  BatchNorm2d(...),
  SiLU(...),
  Conv2d(...),
  BatchNorm2d(...),
  SiLU(...),
  Conv2d(...),
  SiLU(...),
  Conv2d(...),
  Conv2d(...),
  BatchNorm2d(...)]
\end{lstlisting}
            \end{tabular}
        \caption{Blocks.}
        \end{subfigure}
        \caption{Inverted Residual block (a) transformed into a list of layers (b).}
        \label{inverted-residual-blocks-standardization}
        \vspace{-3pt}
    \end{figure}
    
Now we describe the standardization algorithm.
The structure of the model makes it possible to consider it as a tree. 
Each logical component defined by the author has children components, which in the lower level are layers. 
Every component is a node in the tree. The algorithm uses a DFS to parse the graph:

\begin{itemize}
    \item[(a)] For every component, the DFS returns its content represented as a list containing layers or lists of layers.
    \item[(b)] For a node consisting of a single layer, the one-element list with this layer is returned.
    To prepare a result for compound components, all children nodes are processed at first.
    \begin{enumerate}
        \item If the depth of the child result is greater than 1 (the returned list contains other lists) or the number of convolutional and linear layers is at most 1 then the content of the child's list is appended to the current list.
        \item Otherwise, the whole child's list is appended as a single element.
    \end{enumerate}
    \item[(c)] Next, the number of blocks and the number of single layers in the list are computed. A~single layer can also be a list containing a single layer. A block is defined as a list containing at least two layers.
    \item[(d)] If the computed number of single layers is greater than the number of blocks and we are not in the root of the model, all elements in the list are unpacked and their contents are appended separately to the resulting list. Otherwise, the list is returned unchanged.
\end{itemize}

\vspace*{0.25cm}
\begin{algorithm}
\small 
\setlength{\tabcolsep}{2pt}

\SetKwProg{Fn}{}{:}{}
\SetKwFunction{DFS}{blocks-standardization}

\KwData{A module $M$, an integer $d$ - current depth in DFS tree (initially set to 0)}
\KwResult{A list $L$ containing layers or lists of layers}

\Fn{\DFS{$M$, $d$}}{
    Let $L$ be an empty list\;
    \If{$M$ is a single layer}{
        $L$\texttt{.append($M$)}\;
        return $L$\;
    }
    \ForEach{$child$ in $M$\texttt{.children()}}{
        $child_{blocks} = $blocks-standardization($child$, $d$+1)\;
        \If{depth($child_{blocks}$) > 1 \OR number of convolutional + number of linear layers in $child_{blocks}$ $\le$ 1}{
            \ForEach{$block$ in $child_{blocks}$}{
                $L$\texttt{.append($blocks$)}\;
            }
        }\Else{
            $L$\texttt{.append($child_{blocks}$)}\;
        }
    }
    $n_{blocks} \longleftarrow 0$\;
    $n_{layers} \longleftarrow 0$\;
    \ForEach{$block$ in $L$}{
        \uIf{number of layers in $block$ > 1}{
            $n_{blocks} \longleftarrow n_{blocks}+1$\;
        }\Else{
            $n_{layers} \longleftarrow n_{layers}+1$\;
        }
    }
    \If{$n_{blocks} < n_{layers}$ \AND $d > 0$}{
        $L_{tmp} \longleftarrow L$\;
        $L$\texttt{.clear()}\;
        \ForEach{$block$ in $L_{tmp}$}{
            append each element of $block$ to $L$
        }
    }
    return $L$\;
}
\caption{\small Standardization using DFS. Module denotes an entire model or its component at any depth. Function $depth()$ returns the maximal nesting level in the list.}
\label{alg:blocks-standardization}
 \vspace{-3pt}
\end{algorithm}

\subsection{Transfer} 
\label{appendix:transfer}

We present an exact description of a transfer method below.
Let $T, S$ denote the target's and source's tensors respectively. Their sizes are as follows: $T$'s size: $t_1 \times t_2 \times ... \times t_n$,
$S$'s size: $s_1 \times s_2 \times ... \times s_n$.
To describe a mapping we use a function $g$:
\begin{align*}
    g(j, x, y) =
    \begin{cases}
        j + \left \lfloor{\frac{x-y}{2}}\right \rfloor & \mbox{ if }x > y\\
        j & \mbox{ if }x \le y
    \end{cases}
\end{align*}

Then the parameters are mapped this way:
\begin{align*}
    &T_{g(j_1, t_1, s_1), g(j_2, t_2, s_2), ..., g(j_n, t_n, s_n)} = S_{g(j_1, s_1, t_1), g(j_2, s_2, t_2), ..., g(j_n, s_n, t_n)}\\
    &\mbox{for every } j \in \{1,...,min(s_1, t_1)\} \times \{1,...,min(s_2,t_2)\} \times ... \times \{1,...,min(s_n, t_n)\}
\end{align*}

\subsection{Complexity analysis of DPIAT}
\label{appendix:complexity}

The order of the DPIAT is $O(n^2m + p_s + p_t)$ where n is the number of layers in the target and m is the number of layers in the source, $p_s$ is the number of parameters in the source and $p_t$ is the number of parameters in the target. The orders of phases are as follows:
\begin{itemize}
\item standardization $O(nh)$, where n is the number of layers in the network being standardized and h is the depth of the tree structure representing the network implementation, h << n and does not grow with the model size.
\item scoring $O(1)$ as it computes the product of ratios of dimensions and the highest number of dimensions in considered layers is 4 (it can be raised to 5 if we considered 3d convolutions)
\item matching $O(n^2m)$ where n is the number of layers in the target and m is the number of layers in the source. The matching part consists of 2 subphases. In the first one, layers inside blocks are matched for every possible pair of blocks. Let’s denote $n_i$ as the number of layers in the $i$-th block of the target and $m_j$ as the number of layers in the $j$-th block of the source. For every of $(n_i+1)(m_j+1)$ cells in a dynamic programming array it is chosen how many target layers the current source layer will be matched to. It is performed through iteration starting from the current target layer and going to earlier layers. For each cell this process takes $O(n_i)$ time. Thus, the complexity of matching inside a single pair of blocks is $O(n_i^2m_j)$. To get scores between all pairs of blocks, dynamic programming for every possible pair of blocks has to be performed, that has a complexity $O(n^2m)$, because $\sum_i \sum_j n_i^2 m_j = \sum_i n_i^2 \sum_j m_j = \sum_i n_i^2 m = m \sum_i n_i^2 \le m (\sum_i n_i)^2 = mn^2$. The same holds true for the second phase where the blocks are matched. Their pairwise scores are known from the first phase. The number of blocks in the target is $\le n$ and in the source $\le m$. Matching is performed in the same way as in the first phase, so its complexity is dependent on the numbers of blocks as was dependent on the number of layers in the first phase. Thus, the complexity of this phase is also $O(n^2m)$, that gives the total complexity of the matching as $O(n^2m)$.
\item transfer $O(p_s + p_t)$ where $p_s$ is the number of parameters in the source and $p_t$ is the number of parameters in the target. In this phase the layers are already matched and only the copying of the parameters has to be performed.
\end{itemize}
In practice, the total number of layers is a few hundred. For networks, we consider it is equal to: for EfficientNet-B0 - 199, for EfficientNet-B2 - 281, for MixNet-L - 291, for ReXNet - 183. For larger models like EfficientNet-B8 it is equal to 733. If the source and target are of similar size and $n = O(m)$, then the complexity of the DPIAT is $O(n^3 + p_s + p_t)$. For a number of layers smaller than 1000, the matching can be computed within a few seconds. Transfer of the parameters can be performed in parallel on a GPU or on a CPU. The average time of DPIAT between a pair of models is 0.9 s (across pairs presented in Table \ref{tab:cifar100}) and the maximum time is 1.9 s. Compared to that, a single epoch of training of the EfficientNet-B0 on CIFAR100 using Tesla P100 GPU is 200 s.

\section{Additional Experiments and Details}
\label{appendix:exp}

\subsection{Convergence speed up}
\label{appendix:speed-up}


\input{src/results/times}

\subsection{Training on CIFAR100: additional metrics}
\label{appendix:train-on-cifar100}

\begin{figure}[H]
\small 
\setlength{\tabcolsep}{2pt}
\centering
    \includegraphics[width=1.0\textwidth]{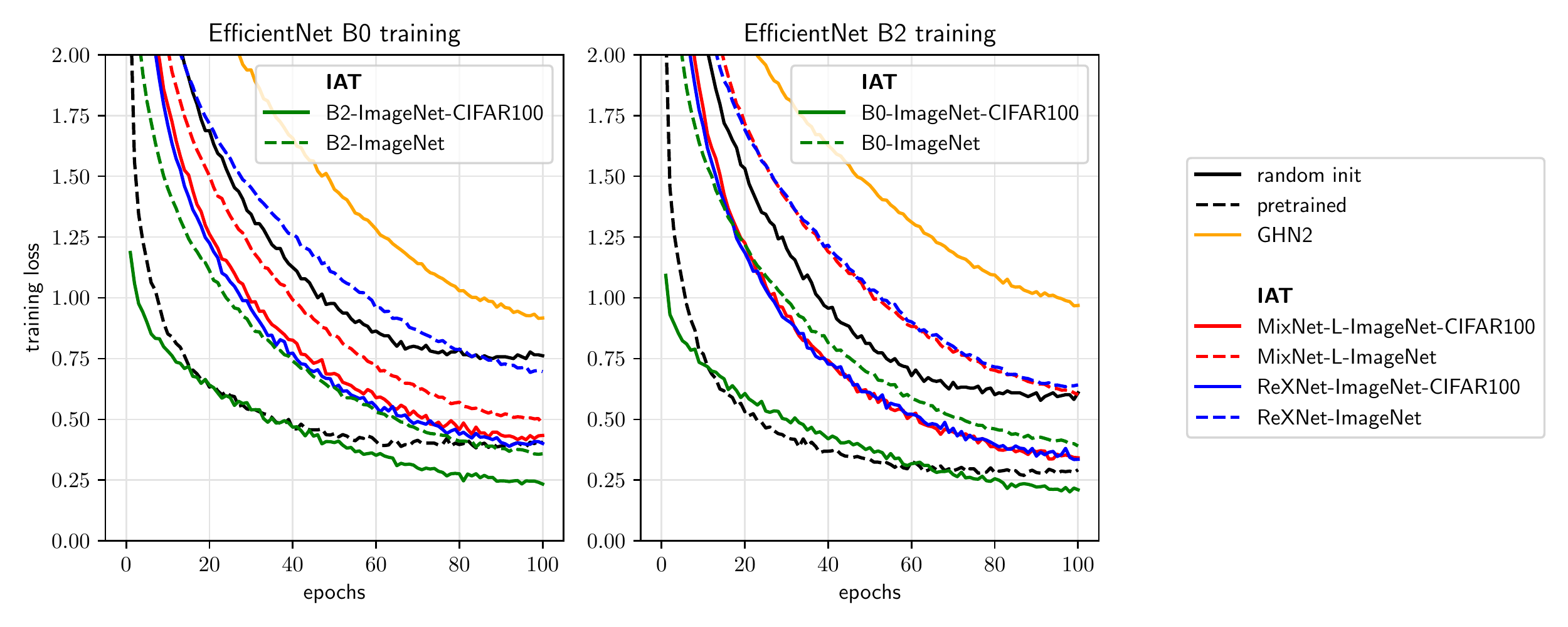}
    \caption{\small Training loss on CIFAR100.}
    \label{fig:cifar100_loss_train}
\vspace{-3pt}
\end{figure}

\begin{figure}[H]
\small 
\setlength{\tabcolsep}{2pt}
\centering
    \includegraphics[width=1.0\textwidth]{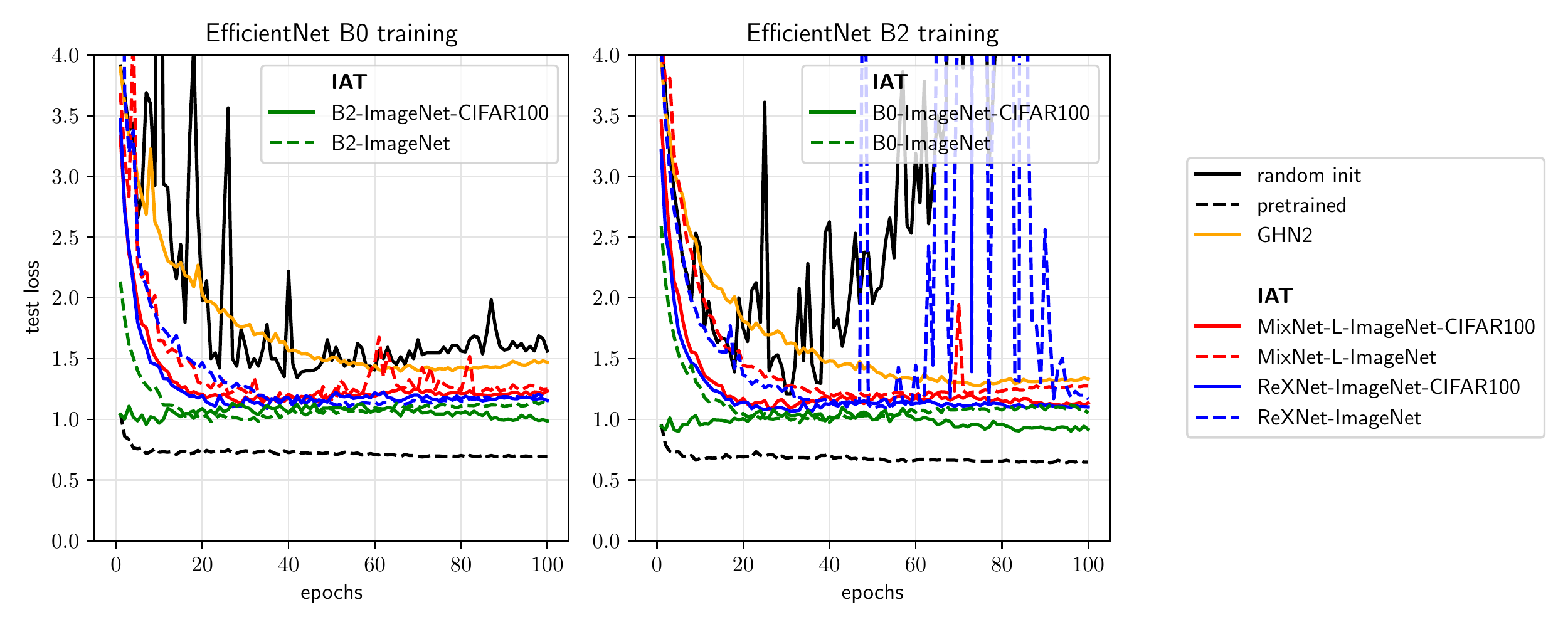}
    \caption{\small Test loss in CIFAR100.}
    \label{fig:cifar100_loss_val}
\vspace{-3pt}
\end{figure}

\begin{figure}[H]
\small 
\setlength{\tabcolsep}{2pt}
\centering
    \includegraphics[width=1.0\textwidth]{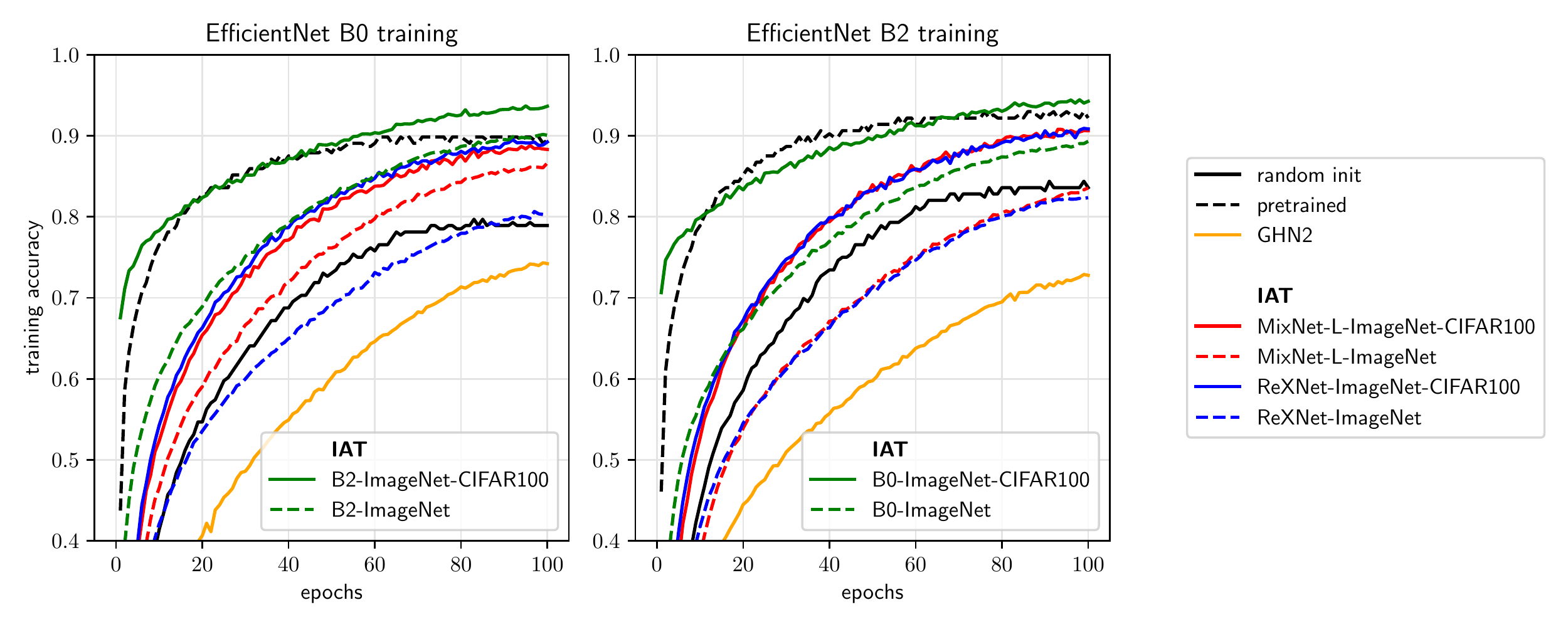}
    \caption{\small Training accuracy on CIFAR100.}
    \label{fig:cifar100_acc_train}
\vspace{-3pt}
\end{figure}

\subsection{Training on ImageNet: additional metrics}
\label{appendix:train-on-imagenet}

\begin{figure}[H]
\small 
\setlength{\tabcolsep}{2pt}
\centering
\minipage{0.33\textwidth}
  \includegraphics[width=1\linewidth]{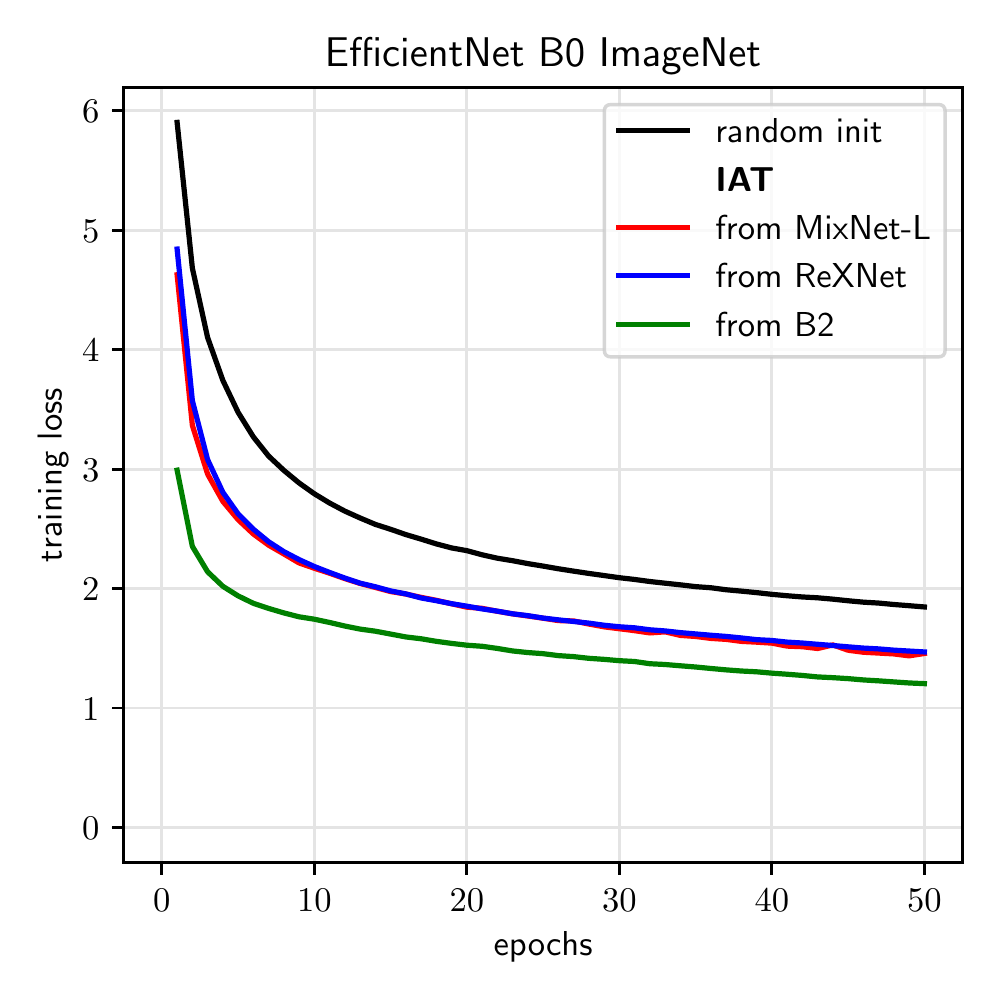}
    \caption{\small Training loss.}
    \label{fig:imagenet_loss_train}
\endminipage\hfill
\minipage{0.33\textwidth}
  \includegraphics[width=1\linewidth]{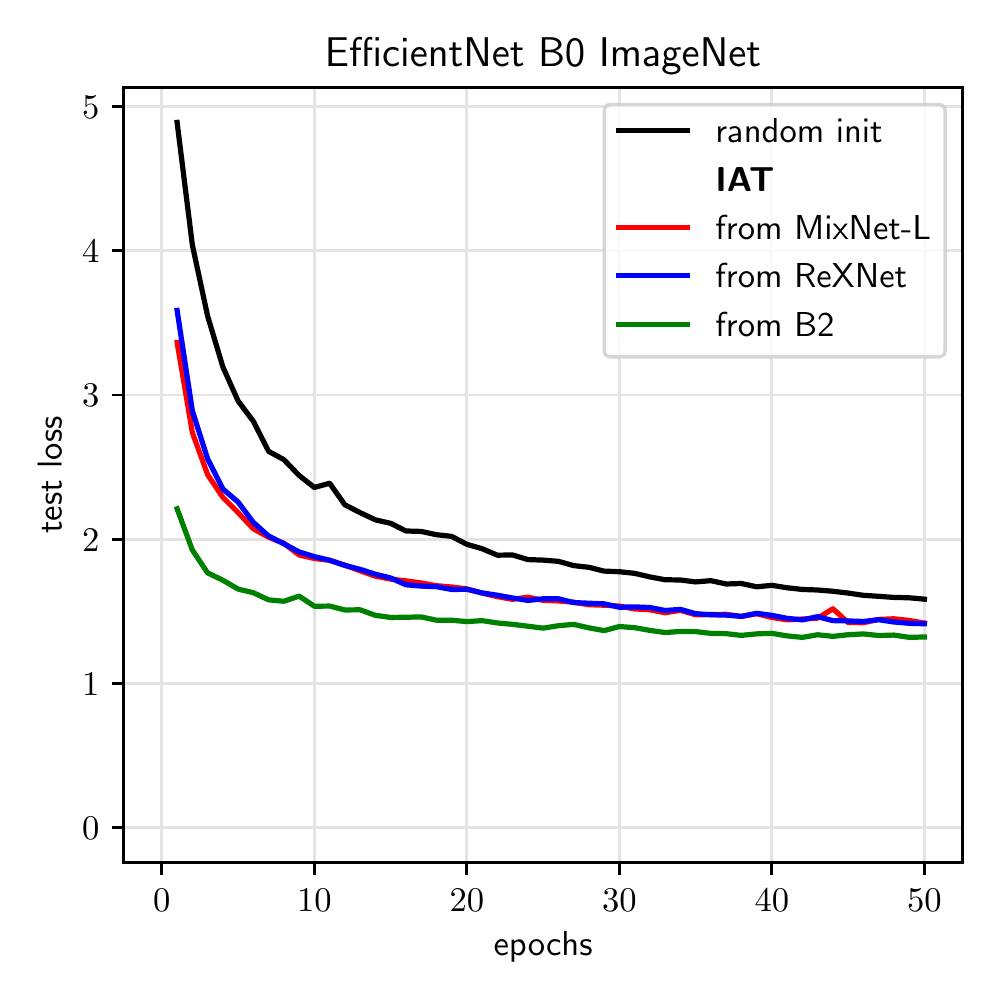}
    \caption{\small Test loss.}
    \label{fig:imagenet_loss_val}
\endminipage\hfill
\minipage{0.33\textwidth}%
  \includegraphics[width=1\linewidth]{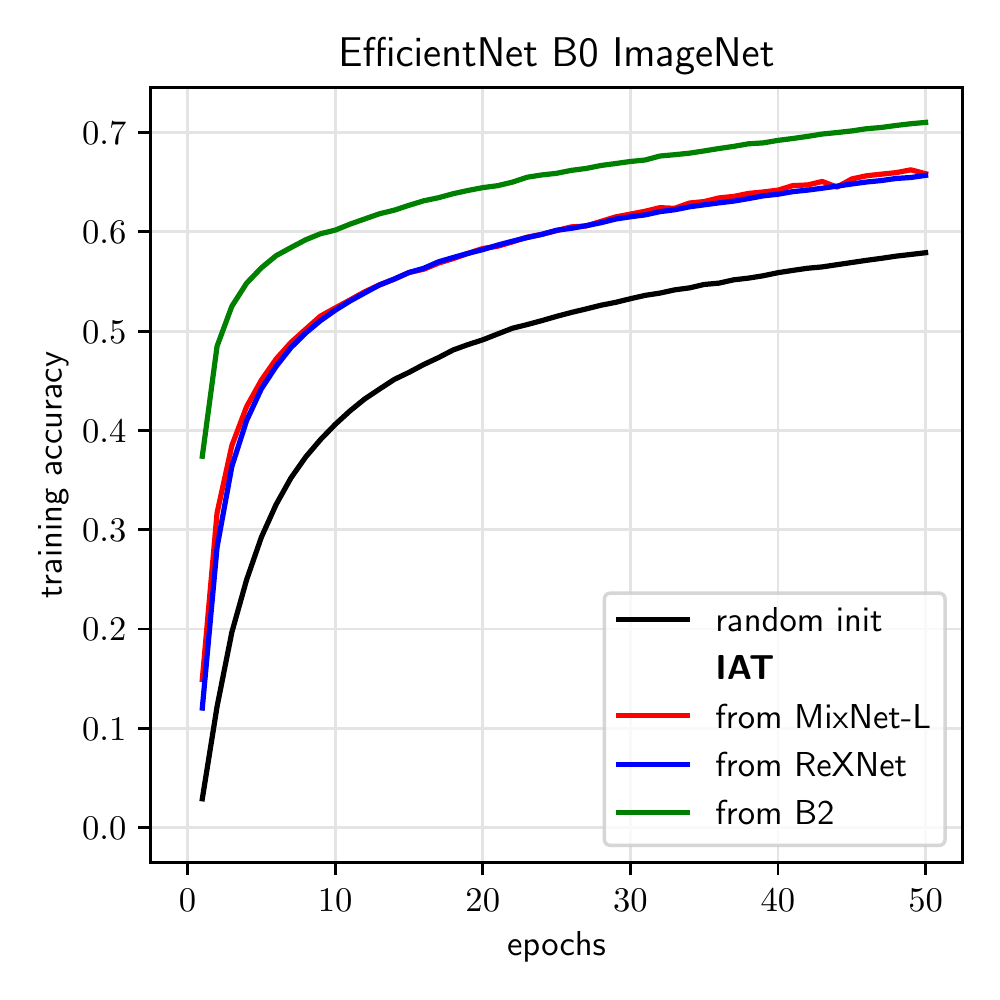}
    \caption{\small Training accuracy.}
    \label{fig:imagenet_acc_train}
\endminipage
\vspace{-3pt}
\end{figure}

%
%

\subsection{More variants on CIFAR100}
\begin{figure}[H]
\small 
\setlength{\tabcolsep}{2pt}
    \centering
    \includegraphics[width=1.0\textwidth]{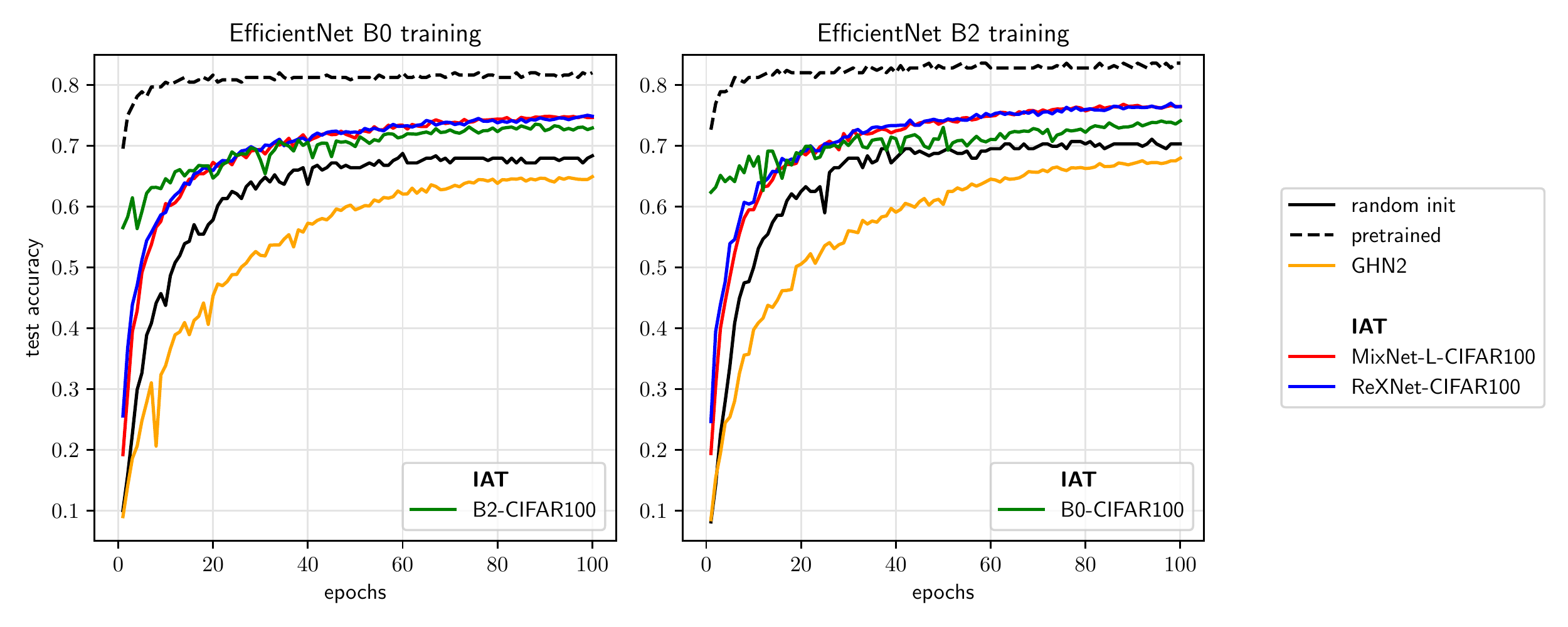}
    \caption{\small
        Test accuracy during CIFAR100 training.
        Black solid line shows accuracy of randomly initialized models.
        A transfer from source model B2-CIFAR100 to EfficientNet-B0 (green solid line) indicates that B2 was randomly initialized and fine-tuned on domain dataset (100 epochs), then used IAT to transfer to B0 and trained for next 100 epochs (this figure).
    } 
    
    \label{fig:cifar100_random_init}
     \vspace{-3pt}
\end{figure}

\subsection{The effectiveness after 4 epochs on more datasets}
\label{appendix:more}
\input{src/results/cifar10}
\input{src/results/food101}

\subsection{ImageNet pretraining}
\label{appendix:imagenetpretraining}
In these experiments we trained 7 models initialized with weights from ImageNet pretraining: EfficientNet-B0, B1, B2, MixNet-M, L MnasNet and SeMnasNet. The training took $100$ epochs.
We then transferred the knowledge to randomly initialized target models and compared test accuracy after 4 epochs with the test accuracy of ImageNet pretrained models after 4 epochs. Exact results for every pair of architectures are available in tables \ref{tab:cifar10_pretrained}, \ref{tab:cifar100_pretrained}, \ref{tab:food101_pretrained}.

\input{src/results/cifar10_pretrained}
\input{src/results/cifar100_pretrained}
\input{src/results/food101_pretrained}

\subsection{Correlation between effectiveness and similarity}
\label{appendix:corr}
\input{src/results/correlation}

\begin{figure}[H]
\small 
\setlength{\tabcolsep}{2pt}
    \centering
    \includegraphics[width=0.95\textwidth]{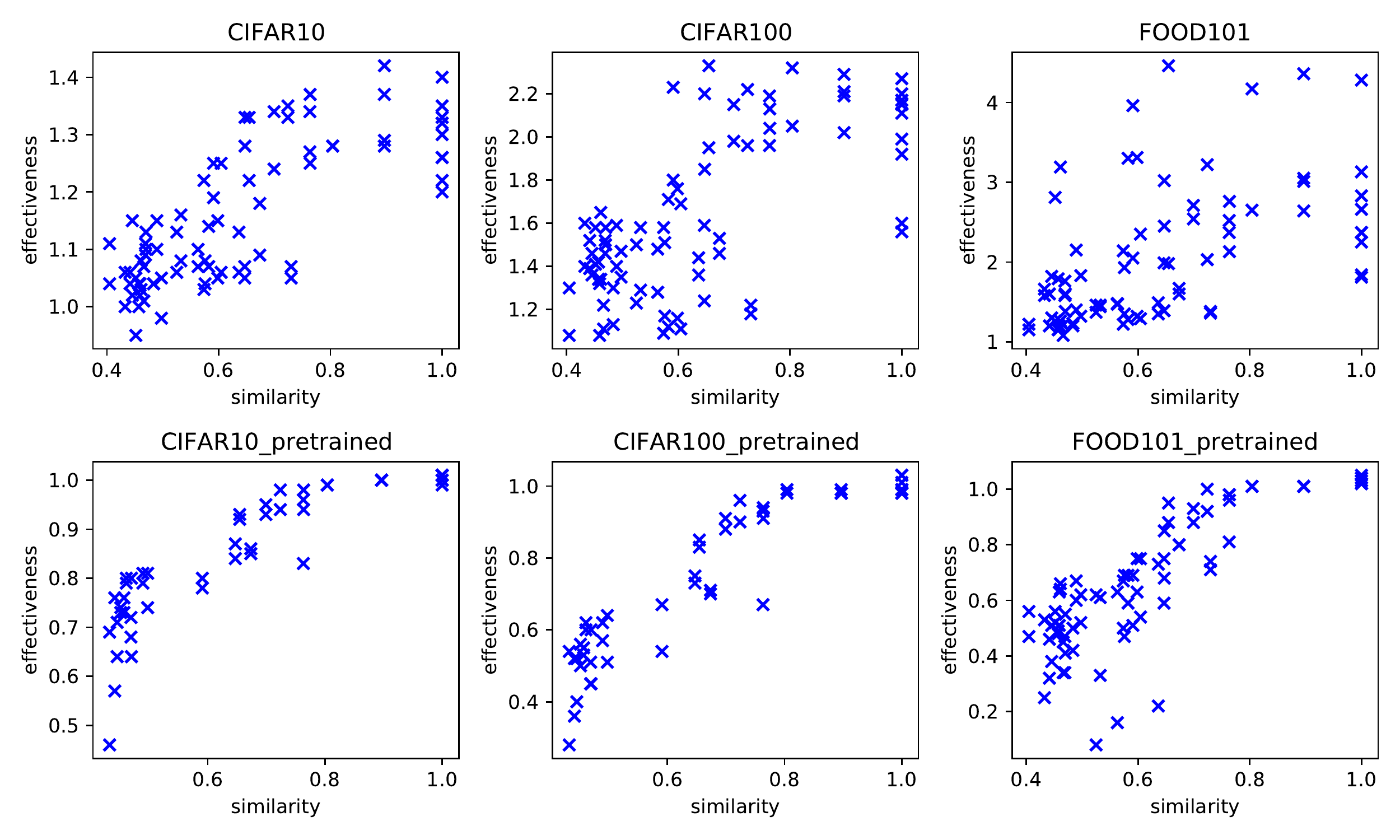}
    \caption{\small Relationship between method effectiveness and architectures similarity. Increase in similarity usually leads to increase in \algo{} effectiveness, but the relationship is not always linear.}
    \label{fig:my_label}
     \vspace{-3pt}
\end{figure}

\subsection{Ablation studies}
\label{appendix:ablation}


Table \ref{tab:methods} shows how our method performed with matching and transfer operators replaced. 

\begin{itemize}
    \item ClipTransfer - copies the center crop of S to the center crop of T
    \item ClipWithNormalization - copies the center crop of S to the center crop of T and applies normalization such that outputs of the consecutive layers have similar variances
    \item FullTransfer - the same as clip transfer when the source dimension is larger - in the opposite case, it replicates the weights from the source to fill target
    \item MagnitudeTransfer - this method copies the connections between layers with the highest magnitude
\end{itemize}

\begin{itemize}
    \item RandomMatching - completely random matching
    \item DPMatching - the main technique described in the paper
    \item BipartiteMatching - relies on DPMatching to match blocks and finds the maximum weight bipartite matching between layers \cite{galil1986}
    \item NBipartiteMatching - like BipartiteMatching, but also uses the same bipartite matching algorithm for blocks
\end{itemize}

\input{src/results/methods}

\section{Frequently Asked Questions}

\subsection{How should results from experiments be interpreted with use-cases scenarios?}
\label{appendix:faq-interpret-results}

\textbf{Initialization problem}. Let's assume we created a novel network. As the network is new, it has never been pretrained on ImageNet. We could pretrain it on ImageNet ourselves, but it has a huge computational cost. In Figure \ref{fig:cifar100} there is a case where MixNet-L is trained on ImageNet and then the parameters are transferred using DPIAT to EfficientNetB0 and fine-tuned on CIFAR100 (Figure \ref{fig:cifar100}, EfficientNet B0 section, MixNet-L-ImageNet). Let's denote this as MixNet-L-ImageNet $\rightarrow$ EfficientNetB0-CIFAR100, and generalize to $N_1$-$D_1$ $\rightarrow$ $N_2$-$D_2$. Imagine that $D_2$ is a completely new dataset, for a problem not known before and we constructed a specifically designed architecture $N_2$. Before IAT, the only available option was to train from scratch. With DPIAT we can do better. In such case, we replace random initialization with DPIAT as follows:
\begin{itemize}
\item Download a pool of pretrained models (e.g.\ from timm library).
\item Use our similarity score to find $N_1$ - the most similar architecture to $N_2$.
\item Use IAT to transfer weights from $N_1$ pretrained on ImageNet to $N_2$ for training on $D_2$.
\item Run training.
\end{itemize}
This results in improvement over random initialization as presented in Figure \ref{fig:cifar100}. If $\similarity(N_1, N_2) \approx 0.8$ it would behave like \textbf{green dashed lines}; If $\similarity(N_1, N_2) \approx 0.5$, it would behave like \textbf{red/blue dashed lines}.

\textbf{Iterative experimentation}. In this scenario many models are trained on a single dataset $D_2$ to improve the scores. This situation applies to Kaggle-like contests where many models are tested to obtain the best score. Let's assume we have $N_2$ that was trained for 100 epochs and we want to test a different model $N_3$ that, for example accepts higher resolution images, has some additional layers and modified shapes of parameter tensors. One option is to initialize the network randomly, but it leads to starting from scratch. Another option is to match layers manually and copy the parameters using Net2Net like methods. This process provides a better initialization point, but is time-consuming and tedious, especially if repeated many times. The third approach is to use DPIAT:
\begin{itemize}
\item Use DPIAT to transfer weights from $N_2$ pretrained on $D_2$ to $N_3$ for $D_2$.
\item Run training.
\end{itemize}
It should train faster than from scratch (Appendix \ref{appendix:speed-up}). It is expected that $N_2$ will give better accuracy if changes to the architecture improve the expressivity of the model. If changes were not significant, both models should stop at the same plateau - similar behavior is presented in Figure \ref{fig:cifar100}. If $\similarity(N_2, N_3) \approx 0.8$ it would behave like \textbf{green solid lines}; If $\similarity(N_2, N_3) \approx 0.5$, it would behave like \textbf{red/blue solid lines}.

This process can be repeated for new models.

\subsection{Why is there no comparison to Net2Net or FNA-like parameter remapping?}
\label{appendix:faq-parameter-remapping}

In the Net2Net paper, the authors propose two operators Net2WiderNet and Net2DeeperNet that allow to transfer knowledge to wider versions of the current layers or add a new layer (initialized as an identity). They admit that their method allows only for transfer to wider and deeper networks. A new network has to be designed based on the current architecture (by widening layers and adding channels). That restricts how different the target and source architectures can be. Our method does not have this limitation.

In FNA++ authors derive new networks from the super network created by the expansion of the seed network. Seed network is an ImageNet pretrained network. Transfer of the parameters is conducted from the seed network to the super network using parameter remapping. However, the architecture of the super network is based on the architecture of the seed network and the remapping process is straightforward. Then, the remapping is performed from the super network to the final network, but the final network is a subnetwork of the super network, so remapping in this case is just a selection of the layers.

Neither Net2Net, nor FNA++ can remap parameters between independently designed architectures. They require that the target network is created from the source network. Our method does not have this limitation. For example, it allows for a transfer from MixNet-L to EfficientNet-B2, without the manual analysis of the architectures, while Net2Net and FNA++ do not.

\input{src/results/comparison}

\section{Limitations}
\label{appendix:limits}
\algo{} reduces the computational burden of iterative optimization methods in a significant way. However, it is still limited in a few aspects.

\textbf{Source network requirement.}
IAT requires a pretrained source network to be available.
Overall, IAT has similar requirements to the regular transfer learning \cite{NIPS2014_375c7134}. 
Consequently, the source network needs to be pretrained on a similar domain to provide with useful features. 

\textbf{Branching.}
While \algo{} can handle multiple branches in a network, multi-branch architecture transfer performance can benefit from advanced network alignment techniques. 

\textbf{Lack of data-driven design.} \cite{fb2021ppuda} could be used to mitigate architecture constraints on the source domain. However, the optimization task is very complicated and resource intensive. Fine-tuning  GHN-2's predicted parameters shows no promising results. Nonetheless, the hypernetwork family is data-driven unlike our method, meaning it has the potential to benefit from large datasets. Whether the parameter prediction task (with the intent to fine tune parameters) is feasible is yet to be shown. 

\textbf{Experimental domain.}
We have not tested the algorithm on problem domains outside CNN-based computer vision (CV) or with the ViT family. In CNN-based CV most layers have semantically similar meaning (i.e.\ compute some abstract features given input), however, there are networks where some layers might not be interchangeable. For example, RNNs have layers which are meant to handle the hidden state in a specific way.


\section{Societal Impact}
\label{appendix:impact}

\textbf{Empowering researchers with limited resources.}
\algo{} can easily be used in manual experimentation. In fact, using \algo{} takes two lines of code without any complicated configuration. As a consequence, we allow researchers to contribute to the AI community in a significant way with less resources.

\textbf{Reducing the impact on the environment.}
Training networks using time and resource consuming optimization methods has a negative impact on the environment.
In particular, training from scratch emits large amounts of greenhouse gasses into the atmosphere \cite{green, mlcarbon}.
By making the already available parameters more reusable we intend to decrease the impact modern deep learning has on the environment.

%% file: src/results/times.tex
\begin{table}[tbhp]
\centering
\vspace{-10pt}

\caption{\small Running time required to achieve specified test accuracy using different initialization methods. Experiments for CIFAR100 were measured on Nvidia Tesla P100 GPU. Accuracy thresholds were chosen as the rounded final accuracy achieved by networks trained with random initialization. GHN2 was not able to provide such accuracy in a specified time. Transfer learning using pretrained\textsuperscript{\textdagger} parameters on ImageNet are provided only as theoretical upper-limit. DPIAT (Fine-tuned) means that there was available a source architecture previously fine-tuned on CIFAR100 dataset.\looseness-1}
\vspace{0.2cm}

\scriptsize
\newcommand{\width}{0.135\textwidth}
\setlength{\tabcolsep}{3pt}

    \begin{tabular*}{\textwidth}{@{\extracolsep{\fill}}r*{3}c*{2}c*{2}c}
    \toprule
    \multicolumn{1}{l}{} & 
        \multicolumn{3}{c}{\tabcap{Initialization}} & 
        \multicolumn{2}{c}{\tabcap{DPIAT}} & 
        \multicolumn{2}{c}{\tabcap{DPIAT (Fine-tuned)}} \\ 
    {\tabcap{Target}} & 
random init & pretrained\textsuperscript{\textdagger} & GHN2 &%
(sim $\approx 0.5$) & (sim $\approx 0.8$) &%
(sim $\approx 0.5$) & (sim $\approx 0.8$) \vspace{1pt}\\
\cline{1-4}\cline{5-6}\cline{7-8}\vspace{-4pt}\\
Eff-B0; 0.65 test acc. & 113 min & \textbf{3.4 min} & - & 76 min & 34 min & 55 min & \textbf{3.4 min} \\
Eff-B2; 0.70 test acc. & 445 min & \textbf{7.1 min} & - & 478 min & 134 min & 177 min & \textbf{7.1 min} \\
\midrule
\tabcap{avg. speedup} & 1x & \textbf{47.9x} & - & 1.2x & 3.3x & 2.28x & \textbf{47.9x} \\
    \bottomrule
    \end{tabular*}
    \label{tab:time_to_acc_cifar100}

\vspace{-3pt}
\end{table}

%
%
%
%

\begin{table}[H]

\centering
\vspace{-10pt}

\caption{\small Running time required to achieve specified $0.60$ test accuracy with EfficientNet-B0 using different initialization methods. Experiment for ImageNet were measured on TPU v2. Accuracy threshold was chosen as the rounded final accuracy achieved by a network trained with random initialization.\looseness-1}
\vspace{0.2cm}

\scriptsize
\newcommand{\width}{0.135\textwidth}
\setlength{\tabcolsep}{3pt}

    \begin{tabular}{lrr}
    \toprule
         \tabcap{Method} & \tabcap{Running time} & \tabcap{speedup}\\
    \midrule
         random init & 705 min & 1x\\
         DPIAT (sim $\approx 0.8$) & \textbf{81 min} & \textbf{8.7x}\\
         DPIAT (sim $\approx 0.5$) & 302 min & 2.3x\\
    \bottomrule
    \end{tabular}
    \label{tab:time_to_acc_imagenet}
\end{table}

%% file: src/results/cifar10.tex
\begin{table}[H]
\centering
\vspace{-10pt}

\caption{\small Effectiveness after 4 epochs of training on \textbf{CIFAR10}. Standard deviations of the effectiveness are always less than 0.11 and their median is equal to 0.01.\looseness-1}
\vspace{0.2cm}

\scriptsize
\newcommand{\width}{0.135\textwidth}
\setlength{\tabcolsep}{3pt}

\begin{tabular*}{\textwidth}{ @{\extracolsep{\fill}}r*{9}{c}}
\toprule
\multicolumn{1}{l}{}&\multicolumn{9}{c}{\tabcap{Source}} \\
{\tabcap{Target}} &             Eff-B0 &    Eff-B1 &    Eff-B2 &              Mix-L &              Mix-M &             ReX1.5 &                ReX &               Mnas &    SeMnas \\
\midrule
Eff-B0 &  $ \mathbf{1.32} $ &  $ 1.29 $ &  $ 1.28 $ &           $ 1.00 $ &           $ 0.98 $ &           $ 1.04 $ &           $ 1.05 $ &           $ 1.24 $ &  $ 1.25 $ \\
Eff-B1 &  $ \mathbf{1.42} $ &  $ 1.40 $ &  $ 1.37 $ &           $ 1.15 $ &           $ 1.15 $ &           $ 1.22 $ &           $ 1.25 $ &           $ 1.33 $ &  $ 1.35 $ \\
Eff-B2 &  $ \mathbf{1.28} $ &  $ 1.28 $ &  $ 1.26 $ &           $ 1.05 $ &           $ 1.03 $ &           $ 1.14 $ &           $ 1.15 $ &           $ 1.19 $ &  $ 1.22 $ \\
Mix-L  &           $ 1.04 $ &  $ 1.02 $ &  $ 0.95 $ &  $ \mathbf{1.26} $ &           $ 1.09 $ &           $ 1.02 $ &           $ 1.04 $ &           $ 1.00 $ &  $ 1.06 $ \\
Mix-M  &           $ 1.05 $ &  $ 1.10 $ &  $ 1.08 $ &           $ 1.18 $ &  $ \mathbf{1.30} $ &           $ 1.11 $ &           $ 1.07 $ &           $ 1.11 $ &  $ 1.13 $ \\
ReX1.5 &           $ 1.08 $ &  $ 1.03 $ &  $ 1.07 $ &           $ 1.04 $ &           $ 1.04 $ &  $ \mathbf{1.22} $ &           $ 1.07 $ &           $ 1.06 $ &  $ 1.08 $ \\
ReX    &           $ 1.07 $ &  $ 1.06 $ &  $ 1.05 $ &           $ 1.04 $ &           $ 1.01 $ &           $ 1.05 $ &  $ \mathbf{1.20} $ &           $ 1.07 $ &  $ 1.06 $ \\
Mnas   &           $ 1.34 $ &  $ 1.28 $ &  $ 1.25 $ &           $ 1.06 $ &           $ 1.09 $ &           $ 1.13 $ &           $ 1.10 $ &  $ \mathbf{1.35} $ &  $ 1.27 $ \\
SeMnas &  $ \mathbf{1.37} $ &  $ 1.33 $ &  $ 1.33 $ &           $ 1.04 $ &           $ 1.10 $ &           $ 1.16 $ &           $ 1.13 $ &           $ 1.34 $ &  $ 1.33 $ \\
\bottomrule
\end{tabular*}
\label{tab:cifar10}
\vspace{-3pt}
\end{table}

%% file: src/results/food101.tex
\begin{table}[H]

\centering
\vspace{-10pt}

\caption{\small Effectiveness after 4 epochs of training on \textbf{FOOD101}. Standard deviations of the effectiveness are always less than 0.12 and their median is equal to 0.03.\looseness-1}
\vspace{0.2cm}

\scriptsize
\newcommand{\width}{0.135\textwidth}
\setlength{\tabcolsep}{3pt}

\begin{tabular*}{\textwidth}{ @{\extracolsep{\fill}}r*{9}{c}}
\toprule
\multicolumn{1}{l}{}&\multicolumn{9}{c}{\tabcap{Source}} \\
{\tabcap{Target}} &    Eff-B0 &    Eff-B1 &    Eff-B2 &              Mix-L &              Mix-M &             ReX1.5 &                ReX &               Mnas &             SeMnas \\
\midrule
Eff-B0 &  $ 2.66 $ &  $ 2.64 $ &  $ 2.65 $ &           $ 1.79 $ &           $ 1.83 $ &           $ 1.93 $ &           $ 1.99 $ &           $ 2.71 $ &  $ \mathbf{2.76} $ \\
Eff-B1 &  $ 3.01 $ &  $ 3.13 $ &  $ 3.05 $ &           $ 1.82 $ &           $ 2.15 $ &           $ 2.14 $ &           $ 2.35 $ &           $ 3.02 $ &  $ \mathbf{3.22} $ \\
Eff-B2 &  $ 4.17 $ &  $ 4.36 $ &  $ 4.28 $ &           $ 2.81 $ &           $ 3.19 $ &           $ 3.30 $ &           $ 3.31 $ &           $ 3.96 $ &  $ \mathbf{4.46} $ \\
Mix-L  &  $ 1.15 $ &  $ 1.30 $ &  $ 1.26 $ &  $ \mathbf{2.25} $ &           $ 1.60 $ &           $ 1.16 $ &           $ 1.23 $ &           $ 1.66 $ &           $ 1.60 $ \\
Mix-M  &  $ 1.32 $ &  $ 1.40 $ &  $ 1.26 $ &           $ 1.67 $ &  $ \mathbf{2.37} $ &           $ 1.15 $ &           $ 1.08 $ &           $ 1.58 $ &           $ 1.60 $ \\
ReX1.5 &  $ 1.35 $ &  $ 1.22 $ &  $ 1.28 $ &           $ 1.20 $ &           $ 1.22 $ &  $ \mathbf{1.84} $ &           $ 1.38 $ &           $ 1.46 $ &           $ 1.46 $ \\
ReX    &  $ 1.39 $ &  $ 1.29 $ &  $ 1.32 $ &           $ 1.20 $ &           $ 1.21 $ &           $ 1.36 $ &  $ \mathbf{1.81} $ &           $ 1.47 $ &           $ 1.49 $ \\
Mnas   &  $ 2.54 $ &  $ 2.45 $ &  $ 2.05 $ &           $ 1.58 $ &           $ 1.76 $ &           $ 1.37 $ &           $ 1.48 $ &  $ \mathbf{2.83} $ &           $ 2.52 $ \\
SeMnas &  $ 2.13 $ &  $ 2.03 $ &  $ 1.98 $ &           $ 1.20 $ &           $ 1.38 $ &           $ 1.44 $ &           $ 1.35 $ &           $ 2.37 $ &  $ \mathbf{2.37} $ \\
\bottomrule
\end{tabular*}
\label{tab:food101}
\vspace{-3pt}
\end{table}

%% file: src/results/cifar10_pretrained.tex
\begin{table}[H]

\centering
\vspace{-10pt}

\caption{\small Validation accuracy after 4 epochs of training on \textbf{CIFAR10} of a target network divided by a validation accuracy after 4 epochs of a network with ImageNet pretrained parameters (the same architecture). Target network was obtained this way: ImageNet pretrained source network $\rightarrow$ 100 epochs of source network training on CIFAR10 $\rightarrow$ \algo{} $\rightarrow$ 4 epochs of target network training on CIFAR10. Standard deviations of the values are always less than 0.11 and their median is equal to 0.02. Transfer between dissimilar architectures give much worse results than network initialized with ImageNet pretrained weights. However, such weights are not always available.\looseness-1}
\vspace{0.2cm}

\scriptsize
\newcommand{\width}{0.135\textwidth}
\setlength{\tabcolsep}{3pt}

\begin{tabular*}{\textwidth}{ @{\extracolsep{\fill}}r*{7}{c}}
\toprule
\multicolumn{1}{l}{}&\multicolumn{7}{c}{\tabcap{Source}} \\
{\tabcap{Target}} &             Eff-B0 &             Eff-B1 &             Eff-B2 &              Mix-L &              Mix-M &               Mnas &             SeMnas \\
\midrule
Eff-B0 &  $ \mathbf{1.01} $ &           $ 1.00 $ &           $ 0.99 $ &           $ 0.73 $ &           $ 0.74 $ &           $ 0.93 $ &           $ 0.94 $ \\
Eff-B1 &           $ 1.00 $ &  $ \mathbf{1.01} $ &           $ 1.00 $ &           $ 0.64 $ &           $ 0.79 $ &           $ 0.84 $ &           $ 0.98 $ \\
Eff-B2 &           $ 0.99 $ &           $ 1.00 $ &  $ \mathbf{1.00} $ &           $ 0.73 $ &           $ 0.80 $ &           $ 0.80 $ &           $ 0.92 $ \\
Mix-L  &           $ 0.76 $ &           $ 0.71 $ &           $ 0.74 $ &  $ \mathbf{0.99} $ &           $ 0.85 $ &           $ 0.69 $ &           $ 0.76 $ \\
Mix-M  &           $ 0.81 $ &           $ 0.81 $ &           $ 0.79 $ &           $ 0.86 $ &  $ \mathbf{1.00} $ &           $ 0.72 $ &           $ 0.80 $ \\
Mnas   &           $ 0.95 $ &           $ 0.87 $ &           $ 0.78 $ &           $ 0.46 $ &           $ 0.68 $ &  $ \mathbf{1.01} $ &           $ 0.83 $ \\
SeMnas &           $ 0.98 $ &           $ 0.94 $ &           $ 0.93 $ &           $ 0.57 $ &           $ 0.64 $ &           $ 0.96 $ &  $ \mathbf{1.01} $ \\
\bottomrule
\end{tabular*}
\label{tab:cifar10_pretrained}
\vspace{-3pt}
\end{table}

%% file: src/results/cifar100_pretrained.tex
\begin{table}[H]

\centering
\vspace{-10pt}

\caption{\small Validation accuracy after 4 epochs of training on \textbf{CIFAR100} of a target network divided by a validation accuracy after 4 epochs of a network with ImageNet pretrained parameters (the same architecture). Target network was obtained this way: ImageNet pretrained source network $\rightarrow$ 100 epochs of source network training on CIFAR100 $\rightarrow$ \algo{} $\rightarrow$ 4 epochs of target network training on CIFAR100. Standard deviations of the effectiveness are always less than 0.24 and their median is equal to 0.03.\looseness-1}
\vspace{0.2cm}

\scriptsize
\newcommand{\width}{0.135\textwidth}
\setlength{\tabcolsep}{3pt}

\begin{tabular*}{\textwidth}{ @{\extracolsep{\fill}}r*{7}{c}}
\toprule
\multicolumn{1}{l}{}&\multicolumn{7}{c}{\tabcap{Source}} \\
{\tabcap{Target}} &    Eff-B0 &             Eff-B1 &             Eff-B2 &              Mix-L &              Mix-M &               Mnas &             SeMnas \\
\midrule
Eff-B0 &  $ 0.98 $ &  $ \mathbf{0.98} $ &           $ 0.98 $ &           $ 0.53 $ &           $ 0.51 $ &           $ 0.88 $ &           $ 0.91 $ \\
Eff-B1 &  $ 0.98 $ &  $ \mathbf{0.99} $ &           $ 0.98 $ &           $ 0.40 $ &           $ 0.57 $ &           $ 0.73 $ &           $ 0.96 $ \\
Eff-B2 &  $ 0.99 $ &           $ 0.99 $ &  $ \mathbf{0.99} $ &           $ 0.50 $ &           $ 0.60 $ &           $ 0.54 $ &           $ 0.83 $ \\
Mix-L  &  $ 0.55 $ &           $ 0.52 $ &           $ 0.56 $ &  $ \mathbf{0.98} $ &           $ 0.71 $ &           $ 0.54 $ &           $ 0.52 $ \\
Mix-M  &  $ 0.64 $ &           $ 0.62 $ &           $ 0.62 $ &           $ 0.70 $ &  $ \mathbf{1.01} $ &           $ 0.51 $ &           $ 0.60 $ \\
Mnas   &  $ 0.91 $ &           $ 0.75 $ &           $ 0.67 $ &           $ 0.28 $ &           $ 0.45 $ &  $ \mathbf{1.03} $ &           $ 0.67 $ \\
SeMnas &  $ 0.94 $ &           $ 0.90 $ &           $ 0.85 $ &           $ 0.36 $ &           $ 0.45 $ &           $ 0.93 $ &  $ \mathbf{1.03} $ \\
\bottomrule
\end{tabular*}
\label{tab:cifar100_pretrained}
\vspace{-3pt}
\end{table}

%% file: src/results/food101_pretrained.tex
\begin{table}[H]

\centering
\vspace{-10pt}

\caption{\small Validation accuracy after 4 epochs of training on \textbf{FOOD101} of a target network divided by a validation accuracy after 4 epochs of a network with ImageNet pretrained parameters (the same architecture). Target network was obtained this way: ImageNet pretrained source network $\rightarrow$ 100 epochs of source network training on FOOD101 $\rightarrow$ \algo{} $\rightarrow$ 4 epochs of target network training on FOOD101. Standard deviations of the effectiveness are always less than 0.24 and their median is equal to 0.02.\looseness-1}
\vspace{0.2cm}

\scriptsize
\newcommand{\width}{0.135\textwidth}
\setlength{\tabcolsep}{3pt}

\begin{tabular*}{\textwidth}{ @{\extracolsep{\fill}}r*{9}{c}}
\toprule
\multicolumn{1}{l}{}&\multicolumn{9}{c}{\tabcap{Source}} \\
{\tabcap{Target}} &             Eff-B0 &             Eff-B1 &             Eff-B2 &              Mix-L &              Mix-M &             ReX1.5 &                ReX &               Mnas &             SeMnas \\
\midrule
Eff-B0 &  $ \mathbf{1.02} $ &           $ 1.01 $ &           $ 1.01 $ &           $ 0.49 $ &           $ 0.52 $ &           $ 0.47 $ &           $ 0.59 $ &           $ 0.88 $ &           $ 0.96 $ \\
Eff-B1 &           $ 1.01 $ &  $ \mathbf{1.02} $ &           $ 1.01 $ &           $ 0.38 $ &           $ 0.60 $ &           $ 0.50 $ &           $ 0.54 $ &           $ 0.68 $ &           $ 1.00 $ \\
Eff-B2 &           $ 1.01 $ &           $ 1.01 $ &  $ \mathbf{1.02} $ &           $ 0.52 $ &           $ 0.64 $ &           $ 0.59 $ &           $ 0.63 $ &           $ 0.51 $ &           $ 0.95 $ \\
Mix-L  &           $ 0.48 $ &           $ 0.51 $ &           $ 0.56 $ &  $ \mathbf{1.02} $ &           $ 0.80 $ &           $ 0.51 $ &           $ 0.42 $ &           $ 0.53 $ &           $ 0.46 $ \\
Mix-M  &           $ 0.62 $ &           $ 0.67 $ &           $ 0.66 $ &           $ 0.80 $ &  $ \mathbf{1.03} $ &           $ 0.47 $ &           $ 0.34 $ &           $ 0.47 $ &           $ 0.55 $ \\
ReX1.5 &           $ 0.69 $ &           $ 0.67 $ &           $ 0.69 $ &           $ 0.63 $ &           $ 0.56 $ &  $ \mathbf{1.03} $ &           $ 0.74 $ &           $ 0.62 $ &           $ 0.61 $ \\
ReX    &           $ 0.75 $ &           $ 0.75 $ &           $ 0.75 $ &           $ 0.50 $ &           $ 0.45 $ &           $ 0.71 $ &  $ \mathbf{1.05} $ &           $ 0.63 $ &           $ 0.73 $ \\
Mnas   &           $ 0.93 $ &           $ 0.85 $ &           $ 0.69 $ &           $ 0.25 $ &           $ 0.34 $ &           $ 0.08 $ &           $ 0.16 $ &  $ \mathbf{1.04} $ &           $ 0.81 $ \\
SeMnas &           $ 0.98 $ &           $ 0.92 $ &           $ 0.88 $ &           $ 0.32 $ &           $ 0.41 $ &           $ 0.33 $ &           $ 0.22 $ &           $ 0.98 $ &  $ \mathbf{1.04} $ \\
\bottomrule
\end{tabular*}
\label{tab:food101_pretrained}
\vspace{-3pt}
\end{table}

%% file: src/results/correlation.tex
\begin{table}[H]
\centering
\vspace{-10pt}

\caption{\small Correlation between method effectiveness and similarity between architectures.\looseness-1}
\vspace{0.2cm}

\scriptsize
\newcommand{\width}{0.135\textwidth}
\setlength{\tabcolsep}{3pt}

    \begin{tabular}{lr}
    \toprule
    \tabcap{Dataset}    &       \tabcap{Correlation}   \\
    \midrule
    CIFAR10             &         0.77 \\
    CIFAR100            &         0.68 \\
    FOOD101             &         0.55 \\
    CIFAR10-pretrained  &         0.88 \\
    CIFAR100-pretrained &         0.92 \\
    FOOD101-pretrained  &         0.83 \\
    \bottomrule
    \end{tabular}
    \label{tab:correlation}
    \vspace{-3pt}
\end{table}

%% file: src/results/methods.tex
\begin{table}[H]

\centering
\vspace{-10pt}

\caption{\small Effectiveness of different methods. It is computed as an average effectiveness over 20 random architectures pairs on CIFAR100. Names are formatted as $X-Y$ where $X$ is the transfer operator and $Y$ is the matching operator.\looseness-1}
\vspace{0.2cm}

\scriptsize
\newcommand{\width}{0.135\textwidth}
\setlength{\tabcolsep}{3pt}
    
    \begin{tabular}{lr}
    \toprule
    \tabcap{Method}                                    & \tabcap{Effectiveness}                  \\
    \midrule
    ClipTransfer-DPMatching                            &  $ 1.59 \pm 0.09$ \\
    ClipTransfer-RandomMatching                        &  $ 0.64 \pm 0.03$ \\
    ClipWithNormalizationTransfer-DPMatching           &  $ 1.05 \pm 0.17$ \\
    FullTransfer-DPMatching                            &  $ 1.59 \pm 0.09$ \\
    MagnitudeTransfer-DPMatching                       &  $ 1.55 \pm 0.09$ \\
    ClipTransfer-BipartiteMatching                     &  $ 1.59 \pm 0.09$ \\
    ClipTransfer-NBipartiteMatching                    &  $ 1.53 \pm 0.08$ \\
    \bottomrule
    \end{tabular}
    \label{tab:methods}
\end{table}

%% file: src/results/comparison.tex
\begin{table}[H]
\centering
\vspace{-10pt}

\caption{\small Overview of initialization methods.\looseness-1}
\vspace{0.2cm}

\scriptsize
\newcommand{\width}{0.135\textwidth}
\setlength{\tabcolsep}{3pt}

\begin{tabular*}{\textwidth}{@{\extracolsep{\fill}}l*{4}l*{1}r}
    \toprule
    \textbf{Method} & 
    \textbf{Technique} &
    \textbf{Complexity} &
    \textbf{Source network} &
    \textbf{Target network} \\
    \midrule
    \textbf{DPIAT} &
        DP matching \& transform \& transfer &
        \cmark\vspace{1pt} Fast &
        \cmark\vspace{1pt} \makecell[l]{Defined by user} &
        \cmark Any \\
    \midrule
    GHN & 
        Predict &
        \cmark\vspace{1pt} \makecell[l]{Fast \\ (inference)} &
        N/A &
        \cmark Any \\
    \midrule
    FNA (remapping)  & 
        \makecell[l]{Transform \& transfer} &
        \cmark\vspace{1pt} Fast &
        \cmark\vspace{1pt} Defined by user &
        \xmark\vspace{1pt} \makecell[l]{Target network scaled \\ from the source network} \\
    \midrule
    Net2Net & 
        Transform \& transfer &
        \cmark\vspace{1pt} Fast &
        \cmark\vspace{1pt} Defined by user &
        \xmark\vspace{1pt} \makecell[l]{ (1) Target network scaled \\ from the source network \\ (2) Only to wider or \\ deeper networks} \\
    \midrule
    \makecell[l]{Pretraining \\ on ImageNet} & 
        Transfer &
        \xmark\vspace{1pt} \makecell[l]{Expensive \\ training} &
        \cmark\vspace{1pt} Defined by user &
        \xmark\vspace{1pt} \makecell[l]{ Target network the\\ same as the source network} \\
    \midrule
    Random init. & 
        Sample &
        \cmark\vspace{1pt} Fast &
        N/A &
        \cmark Any \\
    \bottomrule
\end{tabular*}
\label{tab:comparison}
\vspace{-3pt}
\end{table}